\documentclass[10pt,twocolumn,letterpaper]{article}

\usepackage[pagenumbers]{cvpr} % To force page numbers, e.g. for an arXiv version

\usepackage{graphicx}
\usepackage{amsmath}
\usepackage{amssymb}
\usepackage{booktabs}
\usepackage{enumitem}
\usepackage{bm}
\usepackage{cuted}
\usepackage[sectionbib]{bibunits}

\setlist{nosep, leftmargin=*}

\usepackage[pagebackref,breaklinks,colorlinks]{hyperref}
\usepackage{soul}

\usepackage[capitalize]{cleveref}
\crefname{section}{Sec.}{Secs.}
\Crefname{section}{Section}{Sections}
\Crefname{table}{Table}{Tables}
\crefname{table}{Tab.}{Tabs.}

\newcommand{\speech}{\boldsymbol{y}}
\newcommand{\anim}{\boldsymbol{x}}
\newcommand{\speaker}{\boldsymbol{s}}

\renewcommand{\paragraph}[1]{\vspace{0.03cm} \noindent{\bf #1}}

\begin{document}

%\title{Generative Speech-Driven 3D Facial Animation: Benchmarks and Methods}
\title{Probabilistic Speech-Driven 3D Facial Motion Synthesis:\\ New Benchmarks, Methods, and Applications
%: New Methods and Benchmarks
}

\author{Karren D. Yang, Anurag Ranjan, Jen-Hao Rick Chang, Raviteja Vemulapalli, Oncel Tuzel\\
Apple\\
{\tt\small \{karren\_yang, anuragr, jenhao\_chang, r\_vemulapalli, ctuzel\}@apple.com}
}

\maketitle

%%%%%%%%% ABSTRACT
\begin{abstract}
We consider the task of animating 3D facial geometry from speech signal. Existing works are primarily deterministic, focusing on learning a one-to-one mapping from speech signal to 3D face meshes on small datasets with limited speakers. While these models can achieve high-quality lip articulation for speakers in the training set, they are unable to capture the full and diverse distribution of 3D facial motions that accompany speech in the real world. Importantly, the relationship between speech and facial motion is one-to-many, containing both inter-speaker and intra-speaker variations and necessitating a probabilistic approach. In this paper, we identify and address key challenges that have so far limited the development of probabilistic models: lack of datasets and metrics that are suitable for training and evaluating them, as well as the difficulty of designing a model that generates diverse results while remaining faithful to a strong conditioning signal as speech. We first propose large-scale benchmark datasets and metrics suitable for probabilistic modeling. Then, we demonstrate a probabilistic model that achieves both diversity and fidelity to speech, outperforming other methods across the proposed benchmarks. Finally, we showcase useful applications of probabilistic models trained on these large-scale datasets: we can generate diverse speech-driven 3D facial motion that matches unseen speaker styles extracted from reference clips; and our synthetic meshes can be used to improve the performance of downstream audio-visual models. 
  
\end{abstract}

%%%%%%%%% BODY TEXT
\section{Introduction}

Recently, there has been significant research interest in animating 3D faces from speech signals~\cite{cudeiro2019capture,richard2021meshtalk, fan2022faceformer, xing2023codetalker, danvevcek2023emotional} with
%This task can enable immersive interactions (e.g., in AR/VR) and has 
potential applications across immersive interactions, content creation and synthetic data generation.
%across broad industries \cite{}. 
%
%When a person speaks, their speech is reflected not only in their lip articulation, but also in co-articulated motions in other regions of their face, such as the cheeks, eyebrows, and eyes. A fully realistic face animation model should be able to capture the complex motions present in the whole face, going beyond lip synchronization.
%Many existing works propose to learn the relationship between speech and facial motions in a data-driven manner, leveraging latest advancements in deep learning. However, the facial motions synthesized by these approaches are still far from being fully realistic.
%
%
%
Most existing works approach this problem by learning a \emph{deterministic} mapping from speech to 3D face meshes in a data-driven manner \cite{cudeiro2019capture, fan2022faceformer, xing2023codetalker, danvevcek2023emotional}, leveraging advancements in deep learning. These methods are typically optimized on small datasets containing 10-20 speakers \cite{cudeiro2019capture, fanelli20103} and can achieve high-quality lip reconstruction for the speakers in the training dataset \cite{cudeiro2019capture, fan2022faceformer, xing2023codetalker}. %It is also possible to combine the embeddings of individuals in the training dataset to interpolate between their speaking styles \cite{}.
However, these methods fall short of capturing the \emph{one-to-many} relationship between speech and realistic facial motions.

Animating 3D faces from speech is a complex problem.
%The main challenge in speech-driven face animation is modeling the complex \emph{one-to-many} relationship between speech and facial motions. 
For a given speech utterance, there exists a multi-modal distribution of plausible facial motions capturing large variations in speaking style across a population. Even for a single speaker, the conditional distribution of facial motions given speech is %a multi-modal distribution 
multi-modal, capturing intra-speaker variations such as emotions \cite{danvevcek2023emotional} and other paralingustic cues that give nuance to the meaning of the speech. %
%
%Therefore, it is important to model the multi-modal conditional distribution accurately 
%is important 
%for generating diverse and realistic facial motions.
%
%
%The main challenge is in modeling the complex \emph{one-to-many} relationship between speech and facial expression. %For a given speech utterance, there is a distribution of plausible ways to co-articulate facial gestures. This distribution reflects diversity across a population (such as the individual's speaking style \cite{}), but also intra-speaker characteristics (such as emotions \cite{} and the presence of other paralingustic cues that give nuance to the meaning of the speech \cite{}). Being able to model this distribution is important for generative AI applications (being able to generate a variety of facial motions) as well as for practical applications such as training machine learning models on synthetic datasets (where one would want a distribution of facial motions reflecting all of those found in the real population).
%
%However, most of the existing works focus on learning a deterministic mapping between speech and facial motions \cite{}. 
%
%The objective of this work is to model the complex, one-to-many relationship between speech and 3D facial motion, capturing both inter-personal and intra-personal factors of diversity. 
Modeling this complex, one-to-many relationship between speech and 3D facial motion necessitates a \emph{probabilistic} approach, since approximating a multi-modal distribution with a deterministic point estimate leads to predicting the mean \cite{cudeiro2019capture, fan2022faceformer} or a single mode \cite{xing2023codetalker} of the conditional distribution. % However, this poses several challenges.
%to developing non-deterministic models for this task. 
\subsection{Challenges}
%\rv{It may not be a bad idea to make this a subsection and use "Datasets:", "Metrics:", "Modeling:" in bold font at the beginning of the three paragraphs in this subsection that talk about respective issues.}
\vspace{-0.1cm}
\paragraph{Datasets.} Learning this multi-modal distribution poses new challenges for the field of speech-driven 3D facial animation. 
 First is the limitation of existing datasets.  % and metrics. 
Building a useful probabilistic model that captures the wide variety of speech and facial motions requires a large amount of data from many speakers.
However, existing public datasets are small and contain utterances from few speakers~\cite{cudeiro2019capture, fanelli20103}, thus offering limited opportunity for learning diverse 3D facial motions. While a large-scale dataset is used in MeshTalk \cite{richard2021meshtalk}, this dataset is proprietary and not available to the research community. %

\paragraph{Metrics.} The second challenge is the lack of proper evaluation metrics for probabilistic speech-driven facial motion synthesis. Existing works use lip vertex error as the primary metric for evaluating lip synchronization \cite{cudeiro2019capture, richard2021meshtalk}. While lip vertex error is a useful proxy for lip articulation quality, it presumes a one-to-one relationship between speech and lip motion and penalizes realistic variations from the conditional mean. Other metrics such as upper-face dynamics deviation (FDD) have been proposed to measure the variability of the upper face, but they still compare the generated 3D facial motion against an absolute ground truth \cite{xing2023codetalker}. There is a need for metrics that are more suitable for evaluating lip quality and diversity in a probabilistic setting. 

\paragraph{Modeling.} Third, while learning to model the full distribution paves the way for realistic facial motions, it also opens the door to generating samples that are unexpected or even of lower fidelity \cite{razavi2019generating}. As humans are sensitive to facial cues and expressions, it is crucial for facial motions to be constantly in sync with speech, as any inconsistent motions would be glaringly obvious to a viewer. Most existing probabilistic models in other domains do not consider this problem, as their conditioning signals have weaker correlation with the synthesized content. Therefore, there is a need for modeling techniques that can achieve diverse facial motions while maintaining fidelity to the driving speech signal. Ensuring speech synchronization is made more difficult when also considering the need for other conditioning inputs, namely speaking style. Most existing works do not consider these challenges or interactions as they use one-hot speaker encodings and are not intended to generalize to unseen speaking styles. %There is a need for modeling techniques that can achieve diverse facial motions while maintaining fidelity to the driving speech signal.
\subsection{Contributions}
%To address these challenges, we propose a new benchmark dataset and associated metrics to improve evaluation of speech-driven facial animation methods. We propose a new model based on residual-vector quantization and show its effectiveness in modeling facial motion using the proposed metrics and a \ar{large scale?} perceptual study containing ~\ar{y} responses. We detail our contributions below.
%\subsection{Benchmark Dataset}
In this work, we address these challenges with new large-scale datasets, metrics, and modeling techniques for probabilistic speech-driven 3D facial animation. 

%\paragraph{Contributions.} \rv{Given the verbose description of our contributions, may be we should use a separate section or at least a subsection within introduction. I would vote for a separate contributions section with subsections for 'Benchmark...', 'Modeling techniques' and 'Applications'} In this paper, we aim to address these challenges by proposing (i) benchmark datasets and metrics suitable for developing and evaluating probabilistic methods for speech-driven 3D facial animation, and (ii) modeling techniques that enable diverse and faithful speech-driven 3D facial motion synthesis, conditioned on speech signal as well as a speaking style reference. %Finally, we illustrate the utility of non-deterministic modeling from a perceptual standpoint as well as for generating synthetic data for training audio-visual models. 
%Our concrete contributions are as follows.

\paragraph{Datasets.} We propose a novel benchmark dataset for studying probabilistic speech-driven 3D facial motions based on two large-scale paired audio-mesh datasets derived from the VoxCeleb2 \cite{chung2018voxceleb2} video dataset using state-of-the-art monocular face reconstruction methods \cite{feng2021learning, filntisis2023spectre}. %\rv{Should we use "paired audio-mesh datasets" or "audio-mesh datasets" instead of just "mesh datasets"?} 
Our proposed audio-mesh datasets contain thousands of speakers and are orders of magnitude larger than current public benchmarks \cite{cudeiro2019capture,fanelli20103}.
% While noisier than high-quality 3D motion captures, they serve as effective testbeds for developing and benchmarking methods. 

\paragraph{Metrics.} We introduce metrics that are suitable for evaluating probabilistic models. %The current metric used for this task, lip reconstruction error, characterizes the precision of methods but penalizes diversity. 
We propose to quantify how well probabilistic models generate samples close to the ground truth lip motion, allowing a more comprehensive picture of lip articulation quality that takes the diversity of probabilistic models into account.% \rv{This sentence about "model covering" may be a little difficult to understand. See if we can reword it or provide some other detail to make it better.} 
 We also train audio-mesh synchronization models and speaker recognition models to measure other aspects of generative quality, such as synchronization, realism, and diversity. 
 %\rv{I think this should go to the contributions subsection that talks about various contributions in terms of experiments and results.}

\paragraph{Modeling.} We demonstrate a two-stage probabilistic auto-regressive model over residual vector-quantized codes that achieves diverse generation while maintaining robust synchronization with speech. We show that a standard design of a two-stage probabilistic auto-regressive model that is conditioned on both speech signal and a style reference weakens lip synchronization, and propose a design that can match the speaking style of the reference without sacrificing synchronization quality. We also demonstrate simple but effective sampling strategies for trading off diversity for better lip precision and speech synchronization.

\paragraph{Results.} We use our metrics to analyse prominent deterministic (VOCA \cite{cudeiro2019capture}, Faceformer \cite{fan2022faceformer}, CodeTalker \cite{xing2023codetalker}) and non-deterministic methods (MeshTalk \cite{richard2021meshtalk}) on the large-scale datasets. Our approach outperforms these existing methods, demonstrating the potential of probabilistic modeling. 
In perceptual studies, our approach is rated as producing more realistic lip and upper face motion, as well as more capable of capturing inter-speaker diversity (\emph{i.e.,} matching reference clips) compared to deterministic models. Synthetic lip meshes generated from our method can be used to train downstream audio-visual models. On the challenging task of noisy audio-visual speech recognition on LRS3 \cite{afouras2018lrs3}, we improve relative WER by 11.3\% compared to a model that is trained on the ground truth corpus and 47.0\% compared to meshes from a deterministic model.

\section{Related Work}
%Early methods focus primarily on articulating lip motion with speech, for example, by mapping phonemes to visemes \cite{}. While this yields plausible lip synchronization, omits realistic motions in other regions of the face \cite{}.

%Recent methods animate the 3D geometry of the entire face, e.g., a 3D face mesh. Produce oversmoothed results, lack subtle facial gestures that inherent to real speech patterns

\noindent Speech-driven face animation is a highly active field with extensive literature. Existing works can be grouped as follows. First, there are \emph{viseme-based} methods that map the phonetic components of speech to their visual counterparts. Second, there are \emph{video-based} methods that aim to produce convincing outputs in the pixel space. Third and most relevant to our work, there are \emph{3D animation} methods that drive facial motion as represented by 3D facial landmarks or meshes using speech signals. 

Note there is overlap between these groups, in that some of the photorealistic methods also produce intermediate 3D outputs such as facial landmarks or meshes. However, we draw the distinction depending on whether the techniques mainly focus on the 3D facial geometry or on the photo-realistic video quality. 

\paragraph{Viseme-Based Methods.}
Early methods use linguistic observations \cite{massaro201212, xu2013practical, cao2005expressive, taylor2012dynamic} to map from phoneme to viseme sequences. Phonemes are derived directly from text \cite{ezzat1998miketalk,ezzat2000visual,anderson2013expressive} or from speech via acoustic models \cite{verma2003using}. Viseme sequences are subsequently translated to animations by morphing templates \cite{ezzat1998miketalk,ezzat2000visual,kalberer2002speech,kalberer2001face} %HMM models \cite{wang2012high}, active appearance models \cite{anderson2013expressive}, 
or 3D rigged models as in JALI \cite{edwards2016jali}. More recently, deep learning methods have been introduced to learn the mapping function from phonemes to visemes \cite{zhou2018visemenet,taylor2017deep}. %Notably, Zhou et al. \cite{zhou2018visemenet} propose an LSTM model to map from speech to visemes for controlling a 3D rigged model.
While viseme-based methods provide interpretable controls over lip motion, their expressive power is limited; for example, they cannot produce subtle facial gestures in other regions of the face.

\paragraph{Video-Based Methods.} There is extensive literature on synthesizing photorealistic talking heads from speech inputs. Most of these works synthesize 2D talking head videos \cite{chung2017you, chen2019hierarchical,mittal2020animating,guan2023stylesync,zhang2021facial} and cannot easily be extended to 3D. Some recent methods incorporate neural rendering pipelines to synthesize 3D talking heads that can be rendered from different camera angles \cite{guo2021ad}, but they usually require speaker- or scene-specific training. In general, these methods focus on minimizing errors over the pixels of a video, rather than explicitly modeling 3D facial motions.

%Some video-based methods rely on intermediate modules to model changes in 3D facial landmarks \cite{} or 3DMM parameters \cite{} conditioned on speech. However, these operate on lower-dimensional feature spaces than 3D face meshes, and so the modules are typically optimized using regression losses \cite{} or VAE losses that assume a simple isotropic Gaussian distribution over the facial features \cite{}. %Moreover, these works primarily focus in terms of evaluation, it is difficult to disentangle the quality of the speech-to-3D mapping from the quality of the photorealistic synthesis.
%These intermediate modules do not capture full facial geometry conditioned on speech.
\begin{figure*}[t]
\centering
\includegraphics[scale=0.5]{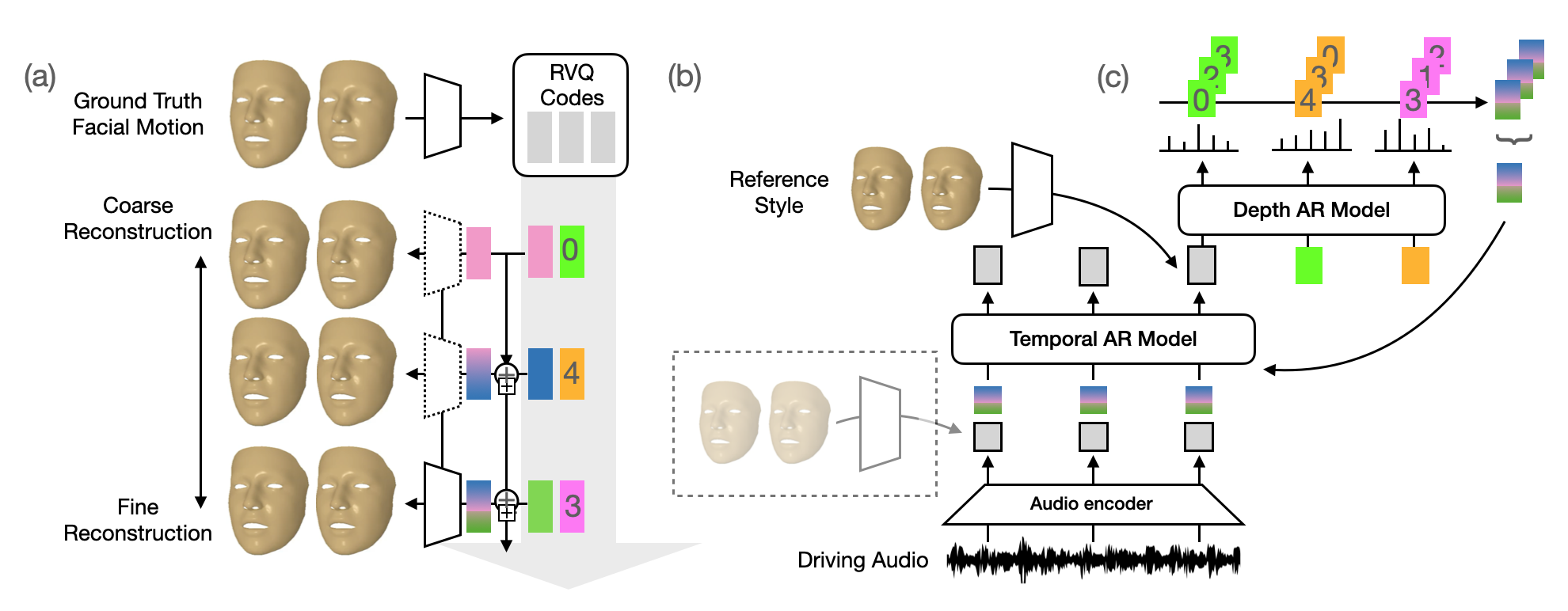}
\caption{\textbf{Method Overview.} We learn a probabilistic model to synthesize 3D facial motion. (a) We first learn a residual vector-quantized codebook over the space of 3D facial motion. (b) We then train a two-stage, probabilistic auto-regressive model to predict these codes in a coarse to fine manner conditioned on audio and a reference speaker clip. (c) During inference, we propose simple and effective sampling strategies to trade-off the diversity of the model in favor of improved lip fidelity. }
\label{fig:method}
\end{figure*}
\paragraph{3D Animation Methods.} Several previous works propose speaker-specific models that need to be trained on personalized data and cannot be used in generic settings \cite{karras2017audio}. Early multi-speaker methods produce low-dimensional features such as blendshape coeffients \cite{deng2006animating}. Recent methods focus on animating the entire face from speech by directly operating in the vertex space \cite{cudeiro2019capture, richard2021meshtalk, fan2022faceformer, xing2023codetalker, danvevcek2023emotional}. However, these methods mostly consider a deterministic formulation of the task. VOCA \cite{cudeiro2019capture} and Faceformer \cite{fan2022faceformer} formulate speech-driven animation as a direct regression problem. In CodeTalker \cite{xing2023codetalker}, the authors explicitly point out this limitation of previous work; even so, their proposed approach does not model the full conditional distribution, but rather projects the output of a regression function to a mode learned from a discrete motion codebook.

%To our knowledge, MeshTalk \cite{} is the only existing work that proposes a probabilistic formulation. However, MeshTalk exhibits weaker lip synchronization than recent deterministic methods and is trained on a large-scale proprietary dataset that is not available to the research community, which has limited follow-up work. Overall, there is a need for benchmark datasets, approaches, and metrics for developing non-deterministic models for speech-driven 3D animation.

%\paragraph{Related Topics} Some works that are related to speech-driven 3D face animation include: (1) dyadic facial motion generation \cite{}, which uses AR modeling to synthesize the listener's facial expressions during conversations; and (2) co-speech gesture synthesis \cite{}, which uses AR modeling to synthesize body gestures that co-occur with speech.

Meshtalk \cite{richard2021meshtalk} proposes a probabilistic method based on learning an auto-regressive model over discrete codes. 
To enforce a correlation between speech and facial motions, lower face vertices are regressed from the speech signal and upper face vertices are reconstructed from ground truth meshes through the bottleneck of a discrete Gumbel-Softmax auto-encoder \cite{richard2021meshtalk}. Subsequently, a probabilistic auto-regressive model is trained over the codes conditioned on speech. While the regression strengthens the correspondence between the speech signal and the generated lip motion, it limits the quality and diversity of the lower face. 

In the context of dyadic 3D facial motion synthesis, Ng \emph{et al.} \cite{ng2022learning} propose a probabilistic auto-regressive model for generating a listener’s facial motions in a two-person conversation. However, the task differs from ours, in that while the listener’s expressions are correlated with the speaker’s voice and motions, this correlation is inherently weaker than in speech-driven facial motion.

\iffalse
\subsection{Generative AR Modeling}
Auto-regressive modeling has demonstrated success in generative tasks across domains such as images \cite{} and audio \cite{}. While initial works applied AR modeling on raw data such as pixels \cite{}, recent methods represent the data as discrete codes and perform AR modeling over these codes. Multiple techniques \cite{} have been developed to learn the discrete codebook. These techniques have been successfully applied to many tasks, including image synthesis \cite{}, audio synthesis \cite{}, text-to-image synthesis \cite{}, text-to-speech synthesis \cite{}, and audio-visual speech enhancement \cite{}.

\paragraph{Discrete representation of facial motion} Several works have used discrete codebooks to represent 3D face meshes. Ng et al. apply vector quantization and discrete AR modeling to generate the listener's facial expressions during conversations \cite{}.

\begin{itemize}
\item Unconditional synthesis
\item Conditional synthesis
\item Vector quantization
\item Most related to our work: Codetalker, learning to Listen: Modeling Non-Deterministic Dyadic Facial Motion. Note: here we are modeling the LISTENER's face, not the speaker's face. It's a different task!
\item Common to capture diversity at the cost of fidelity. Rejection sampling is common, but naive application too slow. Careful architecture and controls are needed.
\end{itemize}
\fi

\section{Approach}
Our goal is to learn a probabilistic model $p_G(\anim | \speech, \speaker)$ to synthesize 3D facial motion from speech, where $\anim \in \mathbb{R}^{T \times 3V}$ is the target sequence of 3D mesh deformations, $\speech \in \mathbb{R}^{T \times D_y}$ is the driving speech signal, and $\speaker \in \mathbb{R}^{T_s \times 3V}$ is a reference speaker sequence of 3D mesh deformations for controlling inter-speaker variation. We propose to first discretize the space of 3D facial motion using a \emph{residual} vector-quantized (RVQ) codebook in a coarse-to-fine manner (Figure \ref{fig:method}a, Section \ref{sec:rvq}). Then, we propose an effective architecture for learning a two-stage probabilistic auto-regressive model over the codes (Figure \ref{fig:method}b, Section \ref{sec:ar-modeling}). %Probabilistic auto-regressive modeling over RVQ codes has been utilized in many other domains \cite{lee2022autoregressive, borsos2023audiolm}, but typically the conditioning signals are not as strongly time-aligned as speech. %and it is advantageous for speech-driven 3D facial animation. As illustrated in Figure \ref{fig:method}(a), RVQ discretizes 3D facial motions in a coarse-to-fine manner, such that facial cues that are synchronized with speech are already present in coarse codes that are predicted first. This improves the synchronization of this method compared to MeshTalk, which learns an auto-regressive model over unordered codes. It also makes it straightforward to trade-off diversity for speed at inference-time, by sampling to a shallower code.
%
%For the two-stage probabilistic auto-regressive model, the standard approach for conditioning the generation on global information (\emph{e.g.,} image class, speaker identity, etc.) is to introduce an embedding into the first AR model, as shown in the dotted line in Figure \ref{fig:method}(b) \cite{lee2022autoregressive}. However, we find this interferes with the strong conditioning signal of the speech. Instead, we propose to introduce the reference speaker information in the depth AR model (Section \ref{sec:ar-modeling}). 
%
Finally, we propose sampling strategies to trade-off diversity for improved precision and speech synchronization, and propose a knowledge distillation strategy to amortize the sampling overhead (Section \ref{sec:trade-off-diversity}).

%First, we train an auto-encoder to map the ground truth animation sequence to a stacked sequence of $D$ RQ codes, representing facial motion in a coarse-to-fine manner (Section \ref{}). The duplex RQ-AR model uses a temporal network to accumulate audio-visual information from previous time points into a context embedding (Section \ref{}) and a separate depth network to predict each of the $D$ codes (Section \ref{}). 

\subsection{RVQ for 3D Facial Motion}\label{sec:rvq}
%\paragraph{Preliminaries: Residual Vector Quantization (RVQ)}
Let $\mathcal{C}$ denote a fixed-size codebook with codes of size $N_C$. Residual vector quantization \cite{lee2022autoregressive} is a discretization technique that recursively projects a vector $\mathbf{z} \in \mathbb{R}^{N_C}$ to the nearest code in $\mathcal{C}$ and takes the residual. After $D$ steps, $\mathbf{z}$ can be represented by an ordered sequence of indices for the codes in $\mathcal{C}$, and the quantization of $\mathbf{z}$ up to depth $d$ is represented by summing the codes corresponding to those indices. 
%Here, $D$ represents maximum code depth of the discretization. To perform a coarser  and the partial sum of the codes in $\mathcal{C}$ indexed by the first $d$ codes is denoted as $\mathbf{z}^{(d)}$.
We apply RVQ to obtain a coarse-to-fine discretization of 3D facial motion by performing the above recursion within the latent space of a 3D facial motion autoencoder, as shown in Figure \ref{fig:method}a. Specifically, we use an temporal convolutional encoder to map $\anim$ to a latent embedding of motion, $Z \in \mathbb{R}^{T \times N_C}$. Each temporal index of $Z$ is separately quantized using RVQ, and the quantized latent embedding of motion is decoded back to the 3D motion space using a convolutional decoder. The encoder and decoder of this autoencoder are jointly optimized via gradient updates to minimize reconstruction loss through the discrete code using a straight-through estimator \cite{van2017neural}. The use of a commitment loss \cite{lee2022autoregressive} to penalize the error of the quantization at every depth effectively ensures that the meshes can be reconstructed from the codes in a coarse-to-fine manner. 

%\paragraph{Discussion} Vector quantization (VQ) has been utilized in related works \cite{} to represent facial motions as discrete codes. The difference between VQ and RQ is the latter represents an embedding as a stack of $D$ codes from a common codebook, and that these codes are ordered to reconstruct the embedding (and original input) in a coarse-to-fine manner. RQ was initially proposed for quantizing images \cite{}, and its advantages include enabling greater expressivity without increasing codebook size.

\subsection{Two-Stage Probabilistic AR Model} \label{sec:ar-modeling}
From RVQ autoencoder, we obtain the codebook indices of a 3D mesh sequence $\anim$. We denote these by a matrix $\textbf{j}$, where $j_{td}$ denotes the index for time point $t$ and depth $d$.
Next, we predict the individual code indices of $\mathbf{j}$ %in an auto-regressive manner, 
conditioned on $\speech$ and $\speaker$,

\begin{equation} \label{eq:AR}
    \prod_{d=1}^D \prod_{t=1}^T p(j_{td} | \mathbf{j}_{<t}, \mathbf{j}_{t, <d}, \speech_{\leq t}, \speaker),
\end{equation}
using a two-stage \cite{lee2022autoregressive} probabilistic Auto Regressive (AR) model consisting of a temporal model and a depth model (Figure \ref{fig:method}b). The temporal model is an auto-regressive model that produces an audio-visual embedding for each time frame $t$ capturing historical audio-visual context as well as the audio embedding from $t$:
\begin{equation}
\mathbf{h}_{av}[t] = \text{TemporalModel}(\speech_{\leq t}, \tilde e(\textbf{j}_{t-1}), \tilde e(\textbf{j}_{t-2}), \cdots)
\end{equation}
where $\tilde e(\textbf{j}_t):= \sum_d e(j_{td})$ and $e(i)$ indicates the code in $\mathcal{C}$ corresponding to the $i$-th index. We experiment with both causal convolutional and transformer auto-regressive architectures for temporal model and find that the longer context of a transformer offers limited benefit when context information is provided through a reference style clip (see Supplemental Materials).
%receives the current and historic context of the audio $\speech_{\leq t}$, as well as embedded codes $\sum_{d=1}^D e(j_{\tilde{t}d})$ for $\tilde{t}<t$, where $e(i)$ denotes the $i$-th code in $\mathcal{C}$, to predict an embedding for the time point $t$. 

\iffalse The temporal model is an auto-regressive network consisting of motion context layers and audio context layers:
\begin{align*}
h_{v}^{l}[t] &=  B_1(h_{av}^{l-1}[1:t-1]), ~\text{for}~1\leq l \leq L \\
h_{av}^{l}[t] &= B_2[h_{v}^{l}[t], a[t]], ~\text{for}~1\leq l \leq L
\end{align*}
where the input $h_{av}^0[t] = \sum_{d=1}^D e(j_{td})$ consists of code embeddings from previous indices and $e(i)$ denotes the $i$-th code in $\mathcal{C}$. 
\fi
%The output of this temporal network is an audio-visual embedding that captures historic and audio context for time frame $t$, denoted by $\mathbf{h}_{av}[t]$. 

Subsequently, the depth model uses the audio-visual context captured in $\mathbf{h}_{av}[t]$ to generate each of the $D$ code indices for the current time frame in an auto-regressive manner. The depth model consists of a masked self-attention transformer block which, at time frame $t$, operates along a length $D+1$ sequence $v_t$ defined as: $v_{t1} = p_1 + E_s(\speaker)$, $v_{t2} = p_2 + \mathbf{h}_{av}[t]$, and $v_{td} = p_d + \sum_{d'=1}^{d-1} e(j_{td'})$ for $d\geq3$, where $p_i$ denotes a learned positional encoding. 
\iffalse
\begin{align*}
v_{t1} &= PE_D(1) + E_s(\speaker) \\
v_{t2} &= PE_D(2) + \mathbf{h}_{av}[t], \\
v_{td} &= PE_D(d) + \sum_{d'=1}^{d-1} e(j_{td'}), \forall 3\leq d \leq D+1
\end{align*}
where $PE$ denotes a positional encoding, and $E_s$ encodes the reference speaker mesh sequence. Specifically, we have,
\fi
The output of the depth model is a prediction of the conditional distribution of the next token.
\begin{equation}
    p(j_{td} | j_{t, <d}, \mathbf{h}_{av}[t], \speaker) 
    = \text{DepthModel}(v_{t, \leq d+1})
\end{equation}
Notice that we incorporate the encoded $\speaker$ as the first token input into the depth transformer, effectively shifting the standard input sequence by one. We find that incorporating speaker information as an input to the second-stage model, rather than as an input to in the first-stage model, which is more standard \cite{lee2022autoregressive} and is showed as the grayed out box in Figure \ref{fig:method}(b), is crucial for proper speech synchronization. As we show in Table \ref{table:ar-ablation}, incorporating the speaker information into the first stage model rather than the second results in a decrease in synchronization. The two-stage auto-regressive model is trained end-to-end to minimize the cross-entropy loss, $- \mathbb{E}_{td} \log p(j_{td} | \mathbf{j}_{<t}, \mathbf{j}_{t, <d}, \speech_{\leq t}, \speaker)$, in a teacher-forcing manner.

\subsection{Trading off Diversity} \label{sec:trade-off-diversity}
During inference, we can sample from the conditional distribution of facial motions as shown in Equation \ref{eq:AR}. This achieves good results but we also want to control the diversity/variability of the synthesis. In particular, the training loss forces the probabilistic AR to capture the entire training distribution of codes, which is noisy and can result in sampling codes that are less faithful to the conditioning speech during inference. Depending on the application, this may be more or less desirable. For example, for generating synthetic training data for a downstream AV model, we may want to have more variability in the generated samples. On the other hand, for driving a 3D avatar using speech, we may care more about the fidelity of the samples at the expense of diversity. Therefore, we provide some sampling strategies to trade-off diversity for fidelity to the speech signal: (1) KNN-based sampling, (2) code averaging, and (3) rejection sampling using a pre-trained synchronization network. For each strategy, as shown in Figure \ref{fig:method}c, we sample multiple codes and aggregate their embeddings before passing the result as the next input to the temporal model.

\paragraph{KNN-based sampling.} For simplicity of notation, let $e_{t} := \tilde e(\textbf{j}_t)$ denote the sampled and reconstructed quantized embedding for time $t$. We replace the sampled code at time step $t$ with the mean of a local Gaussian approximated from its nearest neighbors on the sampling manifold. Let $\mathcal{E}$ denote a set of $N$ codes sampled at time step $t$. We take the estimate $\hat{e}_t$ to be the mean of the set $\{e \in \mathcal{E}~|~|e - e_t| \leq |\text{KNN}_k(e_t, \mathcal{E}) - e_t|\}$,
where $\text{KNN}_k(e_t, \mathcal{E})$ denotes the $k$-th nearest neighbors of $e_t$ in $\mathcal{E}$. The replacement code $\hat{e}_{t}$ is projected to the discrete codebook. 

\paragraph{Code averaging.} We replace the sampled code $e_t$ with a embedding $\hat{e}_t$ given by the mean of $\mathcal{E}$, a set of $N$ codes sampled at time step $t$. The averaged embedding $\hat{e}_t$ is projected to the discrete codebook.

\paragraph{SyncNet-based Sampling.} Inspired by classifier-based rejection sampling in image synthesis, we propose a simple sampling scheme based on a pretrained synchronization network. Specifically, at each time point $t$, we sample and decode a set of $N$ codes. Each code $e_t$ is decoded by the RVQ autoencoder and scored using a pretrained synchronization network. \\

\noindent While these sampling strategies increase the computational overhead of inference, we can amortize them by distilling the modified sampling distributions into a student network that can be run with no additional cost during inference. We do so by relabeling the code inputs as well as targets of  the depth network by the ones obtained from discretizing the aggregated samples (see Supplemental Materials for details).

\section{Experiments}

% \subsection{Non-Deterministic Benchmark}
\subsection{Benchmark Datasets}
Most of the existing works on speech-driven 3D facial motion synthesis use VOCASet \cite{cudeiro2019capture} and BIWI \cite{fanelli20103} for benchmarking. These datasets are small with a limited number of speakers, and models are often trained and evaluated in a speaker-specific manner on these datasets. Because of their small scale and limited speaker diversity, these datasets do not fully capture the complex relationship between speech and facial motions. 
While MeshTalk \cite{richard2021meshtalk} uses a large, multi-speaker dataset to train their model, their dataset is proprietary and not available for public use. %This makes it difficult to build on and evaluate their approach against other emerging methods.

To address this issue, we introduce two large-scale audio-mesh benchmark datasets. These datasets are created by processing videos from the publicly-available VoxCeleb2 video dataset \cite{chung2018voxceleb2} using two monocular face reconstruction methods: DECA \cite{feng2021learning}, a state-of-the-art method for face reconstruction, and SPECTRE \cite{filntisis2023spectre}, a recent method that holds the state-of-the-art for preserving visual speech information. %Both the VoxCeleb2 dataset and the face reconstruction methods are publicly available to the research community.
These two datasets contain face meshes at different granularity enabling us to assess how well different speech-driven facial motion synthesis methods fare on different types of meshes.
%In particular, the meshes extracted by DECA are less detailed than the SPECTRE meshes, exacerbating the one-to-many problem in learning facial motion from speech. 
Table \ref{table:dataset} shows the statistics of the different datasets. Note that VoxCeleb2 is orders of magnitude larger than the existing benchmark datasets, enabling the development of models that capture speaker diversity reflective of a real-world population.
%\rv{We need to provide details of these datasets here such as number of videos, speakers, and any other statistics}

%For quantitative evaluation, lip reconstruction error is used as the main metric to measure lip synchronization quality. 

%This benchmark is geared towards deterministic modeling, as it contains limited speakers and diversity. Moreover, the use of lip reconstruction error as the primary metric presumes a one-to-one relationship between the speech and lip motion per individual. 
%On the other hand, MeshTalk \cite{} uses a large, multi-speaker dataset to train their non-determinsitic model, but the dataset is proprietary and not available for public use. This makes it difficult to build on and evaluate their approach against other emerging methods. %Recent works have benchmarked MeshTalk on VOCASet and BIWI, but 

\begin{table}[t]
    \setlength{\tabcolsep}{0.3em} % increase space between columns slightly for easier reading
    \newcommand{\indentrow}{\hspace{6pt}}
    
    \centering
    \footnotesize
    \resizebox{1\columnwidth}{!}{  % stretch/shrink table to be exactly columnwidth
        \begin{tabular}{l|cc}
            \toprule
            \textbf{Dataset} & ~~\textbf{\# Mesh Sequences}~ & \textbf{\# Speakers} \\
            \midrule
            VOCASet & 480 & 12\\
            BIWI & 1109 & 14\\
            VoxCeleb2 (Mesh) & $>$1M & 6,112 \\
            \bottomrule   
        \end{tabular}
    }
    \caption{\textbf{Comparsion of Different Benchmark Datasets for Speech-Driven 3D Facial Animation.} Our proposed benchmark datasets of meshes reconstructed from VoxCeleb2 are significantly larger than existing benchmark datasets.}
    \label{table:dataset}
\end{table}

\subsection{Metrics}

%In this section, we describe the various metrics we use to benchmark methods on a large-scale diverse dataset. We begin with metrics that capture the quality and realism of lip articulation.

\paragraph{Lip Vertex Error.}
In existing works, lip vertex error is used as the main proxy for lip articulation quality. This metric is calculated as
\begin{equation}
\ell_{vertex}(x, \hat{x}) := \max_{t, i \in \text{lip}} ||\anim_{ti} - \hat{\anim}_{ti}||_2
\end{equation}
where $x$ is the ground truth mesh, $\hat{x}$ is the synthesized mesh, the maximum is taken over all lip vertices and time frames for a given mesh sequence. However, there is a distribution of possible lip vertex positions for a given individual and utterance, and the ground truth is only one sample from this distribution. Lip vertex error does not reflect that a probabilistic model may correctly capture multiple modes that include the ground truth, but receive a large lip vertex error by sampling a different mode. While the lip vertex error measures the precision of the model, or how close every sample is to the ground truth, a more suitable metric for a probabilistic model may be whether any one of several samples, or their mean, is close to the ground truth, which mitigates the impact of diversity on this metric.
% In other words, while lip vertex error provides a notion of model precision (\emph{i.e.}, how close the generated sample is to a ground truth sample), it does not measure \emph{coverage} or how close the ground truth is to the sampling distribution of a probabilistic model. 
%In particular, for the multimodal sampling distribution of a probabilistic model, it is possible that the ground truth and synthesized example were simply sampled from different modes, but that .
%Benchmarking using lip vertex error as a standalone metric encourages learning the conditional mean and penalizes probabilistic models that capture realistic intra-speaker diversity. 

\paragraph{Coverage Error.} To provide a notion of how close the ground truth is to the sampling distribution of a probabilistic model, we propose to generate a set of samples $\mathcal{S}$ and computing the closest distance to the ground truth:
$$\ell_{cover} := \min_{\hat{x} \in \mathcal{S}} \ell_{vertex}(x, \hat{x}).$$
Intuitively, a probabilistic model with small $\ell_{cover}$ has a mode that is close to the ground truth, even if a generated sample is not. 

\paragraph{Mean Estimate Error.} Finally, we also propose to compute the lip vertex error over the mean of $\mathcal{S}$, \emph{i.e.}, $ \ell_{mean} := \ell_{vertex}(x, \mathbb{E}_\mathcal{\hat S}\hat{x})$, to assess how close the mean of the sampling distribution is to the ground truth. Both coverage error and the error of the mean error better reflect whether a probabilistic model is capable of generating the ground truth lip sequence better than computing error from one random sample. Note that all three lip errors are the same for deterministic methods, as they are only capable of generating the same sample.
%
%We also compute the distance between the ground truth and the closest sample in the generated set, i.e., . 
%For sufficiently large $\mathcal{S}$, this quantity reflects how close the ground truth is to the conditional mean $\mathbb{E}[\anim | \speech, \speaker]$ of the distribution learned by the model, which eliminates the confounding factor of diversity when computing $\ell_{vertex}$. 
%Overall, these metrics give an idea of how well the ground truth lip articulation is covered by the model. 
%Note for deterministic methods, all of these quantities are the same.

%, and fails to penalize over-averaged lip meshes that are out of sync with audio (i.e., a slow-moving mouth that regresses the mean can still achieve a relative lower lip reconstruction error). %For our non-deterministic benchmark, where there is greater variability and many valid lip sequences that correspond to a given speech signal, we also consider other metrics for evaluation. 

\paragraph{SyncNet Score.}
%In image and video domains, synthesis quality is often evaluated in the feature space of a pretrained network that captures meaningful semantic information about the sample. 
While lip vertex errors measure how close generated lip articulations are to the ground truth, they do not reflect whether a particular 3D mesh sequence falls into the possible distribution of facial motions conditioned on a speech utterance. We propose to learn this distribution by training an speech-mesh synchronization network that scores how well a mesh corresponds to a given audio, analogous to the lip synchronization metric used in speech-driven video synthesis \cite{chung2017out}.
Specifically, we pretrain two different synchronization networks to assess the alignment between a mesh sequence and an audio signal. In the first network, a multimodal fusion network is used to merge mesh and audio embeddings along the temporal dimension, and a score is computed from the merged embeddings using a linear layer. In the second network, the score is computed directly through the cosine similarity of the normalized mesh and audio embeddings. Both networks are optimized using an InfoNCE contrastive loss \cite{oord2018representation}, and perform well at detecting temporal as well as semantic alignment between audio and 3D face meshes (see Supplemental Materials).

\paragraph{SyncNet Frechet Distance (SyncNet-FD).} Beyond measuring the quality of speech synchronization, we also want to measure how well the speech-related facial motions generated by a model capture the realism and diversity of the real distribution of such motions. To do so, we compute the Frechet distance between 1000 SyncNet embeddings of real and generated mesh sequences from our two pretrained speech-mesh synchronization networks. %Here, we do the same by computing Frechet distance over the features in our two lip synchronization networks.
%In image and video synthesis, FID \cite{} is often used as a measure of both the realism and the diversity of the generated samples. Specifically, it measures the gap between Gaussian approximations of the distribution of real and generated samples within the feature space of a pretrained image classifier.

\paragraph{Style Cosine Similarity and Rank.} While the above metrics provide different measures of how well models generate 3D facial motions corresponding to speech, we also want to measure how well models are able to replicate the diverse speaking styles within the datasets. To do so, we train a speaking style recognition model based on ArcFace \cite{deng2019arcface}, using 3D facial motions as input (\emph{i.e.,} the deformation of ground truth meshes from the neutral templates). We evaluate how well the models are able to replicate a specific individual's speaking style by computing the cosine similarity between the embeddings of the reference speaker mesh sequences and the generated mesh sequences. We also compute the rank of the similarity relative to the similarity of all the other speakers in the training set. Details of the implementation and performance of the recognition model are provided in the Supplemental Materials.

\paragraph{Style Frechet Distance (Style-FD).} Finally, to assess the diversity of speaking styles produces by the model and how well the distribution matches the speaking styles of the real data, we compute the Frechet Distance between the recognition model embeddings of the real and generated mesh sequences.

\begin{figure*}[t]
    \centering
    \hspace{-3mm}
    \begin{subfigure}[t]{0.33\linewidth}
        \centering
    \includegraphics[height=0.86\linewidth]{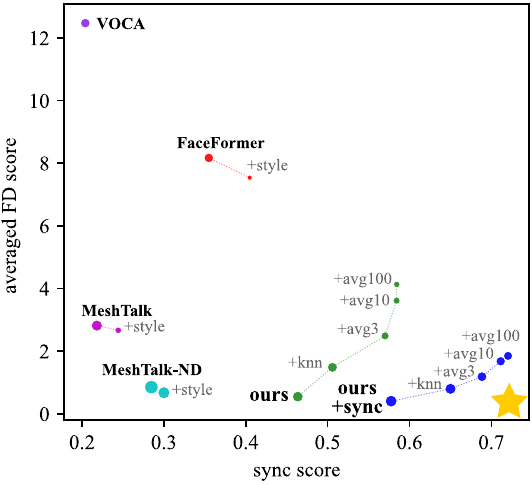}
        \caption{on DECA meshes}
    \end{subfigure}
    \hfill
    \begin{subfigure}[t]{0.33\linewidth}
        \centering
        \includegraphics[height=0.86\linewidth]{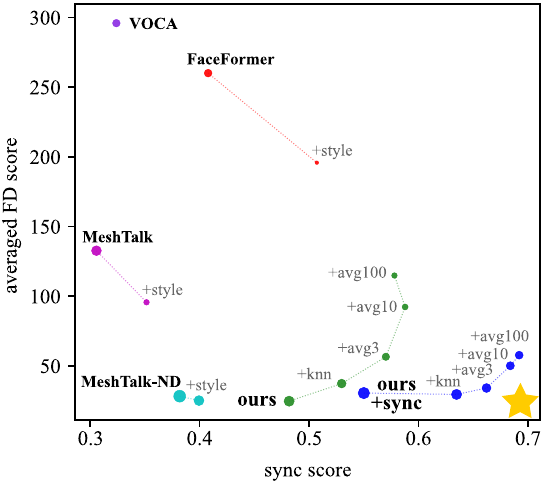}
        \caption{on SPECTRE meshes}
    \end{subfigure}
    \hfill
    \begin{subfigure}[t]{0.33\linewidth}
        \centering
        \includegraphics[height=0.86\linewidth]{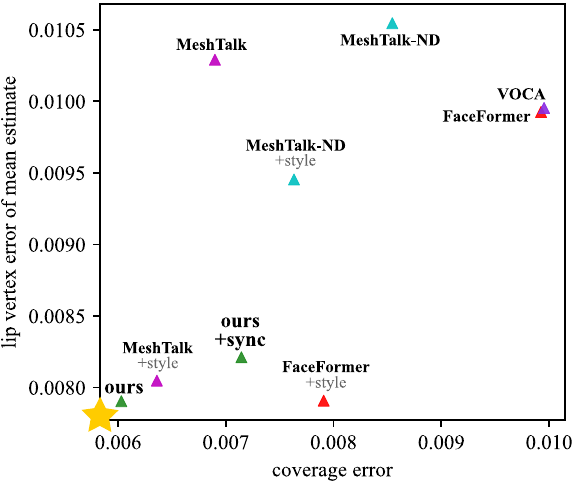}
        \caption{coverage and mean estimate error}
    \end{subfigure}
    % \hfill
    % \begin{subfigure}[t]{0.24\linewidth}
    %     \centering
    %     \includegraphics[width=\linewidth]{cvpr2023-author_kit-v1_1-1/latex/figures/benchmark/spetre_coverage_vs_vertex.pdf}
    %     \caption{}
    % \end{subfigure}    
    \caption{\textbf{Benchmark results.} We evaluate all methods on the aggregate SyncNet score and averaged FD score on (a) DECA and (b) SPECTRE meshes from VoxCeleb2. The size of the dots indicates the lip vertex error. Averaged FD score refers to the average between both SyncNet-FD scores. (c) shows the coverage error and the lip vertex error of the mean estimate over 100 samples per speech input on DECA meshes from VoxCeleb2. The yellow star indicates the direction of the best methods. For averaged FD score, lower is better. For sync score, higher is better. For both coverage and mean estimate error, lower is better. See supplementary material for the complete table.
    }
    \label{fig:benchmark}
\end{figure*}

\subsection{Quantitative Results}
Figure \ref{fig:benchmark} shows the results of our comprehensive benchmark. We train recent deterministic and probabilistic methods on our DECA and SPECTRE meshes and evaluate them using our proposed metrics. Overall, our method outperforms the existing methods on realism/diversity (as measured by averaged FD score), speech synchronization, and lip coverage and mean estimate errors. 
We provide a thorough discussion below.

%For lip vertex error, the circular markers denote $\ell_{vertex}$, while the plus markers and cross markers denote $\ell_{mean}$ and $\ell_{cover}$ respectively. For the SyncNet score and SyncNet-FD, we show the aggregated scores between both SyncNet models. %The full tables can be found in the Supplemental Materials. 
%Overall, our proposed approach outperforms existing methods, producing results with better coverage of ground truth lip articulation, speech synchronization, and realism/diversity. 

\paragraph{Ours \vs Deterministic Methods.}
Existing deterministic methods%
\footnote{We defer discussion of CodeTalker \cite{xing2023codetalker} to the Supplement.} %
suffer in realism/diversity as measured by averaged FD (y-axis, lower is better) on both DECA (Figure \ref{fig:benchmark}a) and SPECTRE meshes (Figure \ref{fig:benchmark}b). Specifically, VOCA \cite{cudeiro2019capture} and Faceformer \cite{fan2022faceformer} are deterministic methods that directly regress 3D mesh vertices on speech, either using a sliding window (VOCA) or an auto-regressive transformer (FaceFormer). Both methods are susceptible to the over-averaging effect of the regression loss, which is exacerbated by training on large-scale datasets. We observe that FaceFormer produces less stiff motions compared to VOCA, due to conditioning on a longer context provided by auto-regressive modeling, resulting in higher synchronization scores. We add conditioning on a reference speaker sequence to FaceFormer to further reduce the distribution of possible facial motions (FaceFormer+Style). This improves its scores across all metrics, particularly on the SPECTRE meshes that are more detailed, but does not resolve the realism/diversity gap.

Our method also outperforms VOCA and Faceformer on speech synchronization, as measured by the sync score in Figures \ref{fig:benchmark}(a-b) (x-axis, higher is better). On the SPECTRE meshes (b), FaceFormer+Style achieves higher SyncNet score compared to default sampling from our probabilistic model, but we can achieve better results using our proposed sampling strategies, as illustrated by the green and blue markers (see later section for discussion). For lip vertex errors, VOCA and FaceFormer both achieve lower $\ell_{vertex}$ (Figure \ref{fig:benchmark}(a-b), marker size, smaller is better), but this is mainly because this metric penalizes the diversity of samples generated by our probabilistic modeling. When we compute the lip vertex error over the average of many samples from our model (Figure \ref{fig:benchmark}(c), y-axis, lower is better) ($\ell_{mean}$, $|\mathcal{S}|=100$), effectively reducing the effect of sampling diversity, we outperform VOCA and FaceFormer and are able to match the lip vertex error of FaceFormer+Style. Furthermore, our model achieves better coverage error than FaceFormer+Style (Figure \ref{fig:benchmark}c, x-axis, lower is better), indicating that our sampling distribution is actually much closer to the ground truth lip vertices. 

\paragraph{Ours \vs MeshTalk.} MeshTalk \cite{richard2021meshtalk} is a two-stage method that first learns a discrete Gumbel-Softmax autoencoder \cite{jang2016categorical} that disentangles upper and lower face motion, then trains a probabilistic auto-regressive model over the discrete codes using a convolutional architecture. While the second stage model is probabilistic, disentangling the lower face involves regressing the vertices from audio over sliding windows, similar to VOCA. We observe that MeshTalk is susceptible to the same over-smoothing effects on the lower face, achieving similar synchronization scores to VOCA in Figure \ref{fig:benchmark}(a-b). Overall, our method achieves better synchronization as well as realism/diversity compared to MeshTalk and MeshTalk+Style, as reflected in higher sync scores and lower averaged FD score. For lip vertex error ($\ell_{vertex}$), our meshes are more diverse, and thus deviate from the ground truth meshes more than MeshTalk+Style. However, we achieve better coverage error as well as mean estimate error, suggesting that while our results are more diverse, our sampling distribution is actually closer to the ground truth. 

We also train a version of MeshTalk without the regression loss in the codebook (MeshTalk-ND, Meshtalk-ND+Style), for a more direct comparison to another probabilistic auto-regressive model that predicts discrete latent codes. Compared to the original version, MeshTalk-ND and MeshTalk-ND+Style are more diverse, as evidenced by lower SyncNet-FD scores, and they are not susceptible to smoothing of the lower face, as evidenced by improved synchronization scores. However, the quality of the lip articulation suffers. Note that MeshTalk-ND+Style cannot cover the ground truth lip sequences as well as MeshTalk+Style or our approach, even though ours is just as diverse. Our approach also achieves higher synchronization scores. This demonstrates the effectiveness of our probabilistic model design choices in maintaining faithfulness to the driving speech signal. 

\paragraph{Trading off Diversity for Fidelity.}
By design, our probabilistic model learns the entire training distribution of RVQ codes, which is noisy and can result in sampling codes that are less faithful to the conditioning speech during inference. The results in Figure \ref{fig:benchmark}(a-b) show that we are able to trade-off diversity for greater fidelity using the strategies in Section \ref{sec:trade-off-diversity} with (blue) or without (green) SyncNet-based rejection sampling. KNN-based sampling achieves a mild trade-off, as code aggregation is based on a local Gaussian approximation. Code averaging achieves a larger trade-off, as the model samples codes that are closer to the conditional mean $\mathbb{E}[\anim_t|\anim_{<t}, \speech, \speaker]$. When averaging between large numbers of codes, eventually the synchronization score decreases due to over-smoothing. 

\paragraph{Speaker Style Evaluation.}
Next, we evaluate the ability of our method to generate the diverse speaking styles of unseen speakers, provided with a reference clip from the target speaker. The results are shown in Table \ref{table:style}, and we compare to other methods that are also trained using a reference clip. Overall, we find that FaceFormer+Style and MeshTalk+Style, which both employ some form of regression from speech in the training stage, are unable to match the speaking style of the target speakers due to over-smoothing and loss of diversity in the facial motions. This is reflected not only in the style cosine similarity, but also in the higher Style-FD. As previous works have noted that recognition networks may be sensitive to slight perturbations introduced by discrete coding schemes \cite{razavi2019generating}, we evaluate our method and MeshTalk-ND+Style on the decompressed ground truth meshes of their respective codebook. We find this improves the style matching scores of both models, which approach the scores of the real ground truth meshes.

\begin{table}[t]
    \setlength{\tabcolsep}{0.3em} % increase space between columns slightly for easier reading
    \newcommand{\indentrow}{\hspace{6pt}}
    
    \centering
    \footnotesize
    \resizebox{1\columnwidth}{!}{  % stretch/shrink table to be exactly columnwidth
        \begin{tabular}{l|ccc}
            \toprule
             \textbf{DECA} & ~~\textbf{Style Cosine Similarity} $\uparrow$~ & \textbf{Style Rank} $\downarrow$ & ~\textbf{Style FD}$\downarrow$~ \\
            \midrule
            FaceFormer+Style & 0.127 &1596.422 & 58.652 \\
            MeshTalk+Style & 0.229 & 1135.0 & 38.535  \\
            MeshTalk-ND+Style  & 0.629 & 53.8 & \textbf{17.068}  \\
            Ours & \textbf{0.707} & \textbf{7.3} & 21.038  \\
            \midrule
            GT & 0.7644 & 10.691 & - \\
            \midrule
             \textbf{SPECTRE} & ~~\textbf{Style Cosine Similarity}$\uparrow$~ & \textbf{Style Rank} $\downarrow$& ~\textbf{Style FD}$\downarrow$~ \\
            \midrule
            FaceFormer+Style & 0.237 & 650.128 & 41.062 \\
            MeshTalk+Style & 0.231 & 955.3 & 69.224 \\
            MeshTalk-ND+Style & 0.609 & 38.8 & \textbf{20.560}  \\
            Ours & \textbf{0.673} & \textbf{20.5} & 23.533   \\
            \midrule
            GT & 0.7522 & 4.982 & - \\
            \bottomrule   
        \end{tabular}
    }
    \caption{\textbf{Style Similarity Scores} show that our probabilistic approach can synthesize facial motion closer to the reference style compared to other deterministic methods. See text for details.}
    \label{table:style}
\end{table}

\paragraph{Key Design Choices for AR Modeling.} One challenge of our task is capturing the diverse facial motions corresponding to speech while maintaining faithfulness to speech signal. In Table \ref{table:ar-ablation}, we show the results of ablation studies that highlight our key design choices. First, using a convolutional architecture for the auto-regressive modeling, as in MeshTalk, results in significantly worse sync scores. Second, incorporating style information early in the temporal AR model, rather than the depth AR model, as done in many works that condition on global embeddings, significantly impairs the synchronization score.

\begin{table}[t]
    \setlength{\tabcolsep}{0.3em} % increase space between columns slightly for easier reading
    \newcommand{\indentrow}{\hspace{6pt}}
    
    \centering
    \footnotesize
    \resizebox{1\columnwidth}{!}{  % stretch/shrink table to be exactly columnwidth
        \begin{tabular}{l|ccc}
            \toprule
            Method & ~~\textbf{Style Cosine Similarity} $\uparrow$~ & \textbf{Sync Score} $\uparrow$ & ~\textbf{Sync FD} $\downarrow$~ \\
            \midrule
            AR-ConvNet (no style) & - & \textcolor{red}{0.217}& \textcolor{red}{9.58}\\
            AR-Transformer (no style) & - & 0.442& 4.21\\
            AR-Transformer+ES  & 0.315 & \textcolor{red}{0.287}& 2.95\\
            Ours & 0.298 & 0.4634 & 3.21 \\
            \bottomrule   
        \end{tabular}
    }
    \caption{\textbf{Key Ablations} of our model. We show that the design of the auto-regressive model is crucial for proper synchronization.}
    \label{table:ar-ablation}
\end{table}

%\subsection{Trading off Diversity for Speed}
\iffalse
\begin{table}[t]
    \setlength{\tabcolsep}{0.3em} % increase space between columns slightly for easier reading
    % \renewcommand{\arraystretch}{1.1}% for the vertical padding
    \newcommand{\indentrow}{\hspace{6pt}}
    
    \centering
    \footnotesize
    \resizebox{1\columnwidth}{!}{  % stretch/shrink table to be exactly columnwidth
        \begin{tabular}{c|ccc}
            \toprule
             ~~\textbf{\#Codes}~~ & Speed ($\times$RT)* & \textbf{Sync Score} & ~\textbf{Sync FD}~ \\
            \midrule
            16 & 1.2 & & \\
            12 & 1.3 & & \\
            10 & 1.5 & & \\
            8 & 1.7 & & \\
            6 & 2.1 & & \\
            4 & 2.3 & & \\
            3 & 2.9 & & \\
            \bottomrule   
        \end{tabular}
    }
    \caption{}
    \label{table:speed}
\end{table}
\fi

\subsection{Applications}

\begin{table}[t]
    \setlength{\tabcolsep}{0.3em} % increase space between columns slightly for easier reading
    \newcommand{\indentrow}{\hspace{6pt}}
    
    \centering
    \footnotesize
    \resizebox{1\columnwidth}{!}{  % stretch/shrink table to be exactly columnwidth
        \begin{tabular}{l|ccc}
            \toprule
             & \textbf{Style Matching} & \textbf{Lip Realism} & \textbf{Upper Face Realism} \\
            \midrule
            Ours \vs VOCA & 85.1/8.5/6.4 & 70.2/17.0/12.8 & 80.9/4.3/14.9 \\
            Ours \vs FaceFormer  & 74.4/20.5/5.1 & 59.0/35.9/5.1 & 71.1/26.3/2.6 \\
            Ours \vs CodeTalker  & 78.8/9.1/12.1 & 87.9/3.0/9.1 & 90.9/3.0/6.1\\
            Ours \vs Faceformer+Style  & 75.0/11.1/13.9  & 86.1/8.3/5.6 & 94.4/2.8/2.8 \\
            \bottomrule   
        \end{tabular}
    }
    \caption{\textbf{Results of a Perceptual Study}. Results show percentage of survey respondents who preferred Ours / Baseline / Neither on each of the categories. For style matching, users were provided a reference clip in addition to two videos and asked which one matched the style in the clip better, which one had more realistic lower lip motion, and which one had more realistic upper face motion.}
    \label{table:perceptual_study}
\end{table}
\begin{table}[t]
    \setlength{\tabcolsep}{0.3em} % increase space between columns slightly for easier reading
    \newcommand{\indentrow}{\hspace{6pt}}
    
    \centering
    \footnotesize
    \resizebox{1\columnwidth}{!}{  % stretch/shrink table to be exactly columnwidth
        \begin{tabular}{ll|c}
            \toprule
            Training Data Type & Training Data Corpus & WER $\downarrow$ \\
            \midrule
            Audio-only & LRS3 trainval+pretrain & 18.7\\
            Real AV & LRS3 trainval & 30.7 \\
            \midrule
            ~~~+ Faceformer Synthetic Dataset & LRS3 pretrain & 13.4\\
            ~~~+ Ours Synthetic Dataset & LRS3 pretrain & \textbf{7.1} \\
            ~~~+ Real AV & LRS3 pretrain & 8.0 \\
            \bottomrule   
        \end{tabular}
    }
    \caption{\textbf{Synthetic Data Generation for AVSR} Training an audio-visual speech recognition model on synthetic meshes generated by our model improves WER over training on meshes extracted from ground truth videos.}

    \label{table:avsr}
\end{table}

We showcase two useful applications of a probabilistic model trained on a diverse large-scale dataset. The first application is the ability to generate more natural and realistic 3D facial motions that capture a diversity of real-world speaking styles, including being able to match the style from a reference clip. We show the results of user ratings in Table \ref{table:perceptual_study}, illustrating that our approach is strongly preferred over prominent deterministic methods trained on smaller, high-quality datasets, as well as FaceFormer+Style trained on our large-scale datasets. Second, we demonstrate the utility of probabilistic methods for generating synthetic training data for downstream audio-visual tasks. Specifically, we consider the challenging task of noisy audio-visual speech recognition (noisy-AVSR) on the Lip Reading Sentences 3 (LRS3) dataset. High-quality synthetic training data is immensely useful for audio-visual speech recognition, not only because labeled audio-visual corpora are limited, but also because there may be privacy concerns with training and deploying a model on real user data. We show that synthetic data from our speech-driven 3D facial animation model can greatly improve the performance of such audio-visual models, even compared to training on the ground truth visual data. We use our model trained on SPECTRE meshes to generate a large, synthetic 3D facial mesh dataset corresponding to the audio in the ``pretrain'' subset of the LRS3 dataset and use the detailed lip meshes as input to the downstream model. As shown in Table \ref{table:avsr}, training an audio-visual speech recognition model on this synthetic visual corpus improves relatively the WER of the model on the test set of LRS3 by 11.3\% compared to training on the ground truth lip meshes, and by 47\% compared to training on meshes generated by FaceFormer (also trained on SPECTRE meshes). Beyond creative applications, this demonstrates the practical usage of non-deterministic 3D facial mesh synthesis methods for training downstream audio-visual models.

\section{Conclusion} In this work, we propose a new large-scale dataset, metrics, and methodology to address the task of probabilistic speech-driven 3D facial motion synthesis. We show the advantages of probabilistic approaches to this task in capturing diversity and propose a careful model design and sampling strategies to ensure strong lip synchrony. We benchmark existing methods on our large-scale dataset, analysing the strengths and weaknesses of each approach. Our probabilistic model outperforms the existing methods across metrics capturing realism, diversity and lip synchronization. Furthermore, we demonstrate new uses of probabilistic models trained on our large-scale dataset for (i) generating 3D facial motion that matches real-world speaking styles through a reference clip, as well as (ii) generating synthetic training data for a downstream audio-visual speech recognition task, which are only possible due to the diversity captured by the dataset and probabilistic model. This work provides a useful large-scale benchmark and analysis tools for other researchers working on this task.

\paragraph{Acknowledgment} We thank Zak Aldeneh, Barry Theobald, Masha Fedzechkina Donaldson, Dianna Yee, and Hadi Pour Ansari for helpful discussions and feedback on the paper.

\iffalse
\paragraph{Quantitative Results - Style}
\begin{itemize}
    \item Our approach is also the only one capable of zero-shot style matching
\item TODO: style continuation scores
\item TODO: style matching scores
\end{itemize}

\paragraph{Subjective Evaluation}

\paragraph{Controlling Variability}

\paragraph{Rejection sampling}

\paragraph{Ablations}
\begin{itemize}
\item What happens when we don't use long-term short-term separation of context? No context vs. style matching vs. overfitting.
\end{itemize}

\paragraph{Qualitative Results - Combination with Decoder}
\fi

%%%%%%%%% REFERENCES
{\small
\bibliographystyle{ieee_fullname}
\bibliography{egbib}

\begin{thebibliography}{10}\itemsep=-1pt

\bibitem{afouras2018lrs3}
Triantafyllos Afouras, Joon~Son Chung, and Andrew Zisserman.
\newblock Lrs3-ted: a large-scale dataset for visual speech recognition.
\newblock {\em arXiv preprint arXiv:1809.00496}, 2018.

\bibitem{anderson2013expressive}
Robert Anderson, Bjorn Stenger, Vincent Wan, and Roberto Cipolla.
\newblock Expressive visual text-to-speech using active appearance models.
\newblock In {\em Proceedings of the IEEE conference on computer vision and
  pattern recognition}, pages 3382--3389, 2013.

\bibitem{baevski2020wav2vec}
Alexei Baevski, Yuhao Zhou, Abdelrahman Mohamed, and Michael Auli.
\newblock wav2vec 2.0: A framework for self-supervised learning of speech
  representations.
\newblock {\em Advances in neural information processing systems},
  33:12449--12460, 2020.

\bibitem{cao2005expressive}
Yong Cao, Wen~C Tien, Petros Faloutsos, and Fr{\'e}d{\'e}ric Pighin.
\newblock Expressive speech-driven facial animation.
\newblock {\em ACM Transactions on Graphics (TOG)}, 24(4):1283--1302, 2005.

\bibitem{chen2019hierarchical}
Lele Chen, Ross~K Maddox, Zhiyao Duan, and Chenliang Xu.
\newblock Hierarchical cross-modal talking face generationwith dynamic
  pixel-wise loss.
\newblock {\em arXiv preprint arXiv:1905.03820}, 2019.

\bibitem{chung2017you}
Joon~Son Chung, Amir Jamaludin, and Andrew Zisserman.
\newblock You said that?
\newblock {\em arXiv preprint arXiv:1705.02966}, 2017.

\bibitem{chung2018voxceleb2}
Joon~Son Chung, Arsha Nagrani, and Andrew Zisserman.
\newblock Voxceleb2: Deep speaker recognition.
\newblock {\em arXiv preprint arXiv:1806.05622}, 2018.

\bibitem{chung2017out}
Joon~Son Chung and Andrew Zisserman.
\newblock Out of time: automated lip sync in the wild.
\newblock In {\em Computer Vision--ACCV 2016 Workshops: ACCV 2016 International
  Workshops, Taipei, Taiwan, November 20-24, 2016, Revised Selected Papers,
  Part II 13}, pages 251--263. Springer, 2017.

\bibitem{cudeiro2019capture}
Daniel Cudeiro, Timo Bolkart, Cassidy Laidlaw, Anurag Ranjan, and Michael~J
  Black.
\newblock Capture, learning, and synthesis of 3d speaking styles.
\newblock In {\em Proceedings of the IEEE/CVF Conference on Computer Vision and
  Pattern Recognition}, pages 10101--10111, 2019.

\bibitem{danvevcek2023emotional}
Radek Dan{\v{e}}{\v{c}}ek, Kiran Chhatre, Shashank Tripathi, Yandong Wen,
  Michael~J Black, and Timo Bolkart.
\newblock Emotional speech-driven animation with content-emotion
  disentanglement.
\newblock {\em arXiv preprint arXiv:2306.08990}, 2023.

\bibitem{deng2019arcface}
Jiankang Deng, Jia Guo, Niannan Xue, and Stefanos Zafeiriou.
\newblock Arcface: Additive angular margin loss for deep face recognition.
\newblock In {\em Proceedings of the IEEE/CVF conference on computer vision and
  pattern recognition}, pages 4690--4699, 2019.

\bibitem{deng2006animating}
Zhigang Deng, Pei-Ying Chiang, Pamela Fox, and Ulrich Neumann.
\newblock Animating blendshape faces by cross-mapping motion capture data.
\newblock In {\em Proceedings of the 2006 symposium on Interactive 3D graphics
  and games}, pages 43--48, 2006.

\bibitem{edwards2016jali}
Pif Edwards, Chris Landreth, Eugene Fiume, and Karan Singh.
\newblock Jali: an animator-centric viseme model for expressive lip
  synchronization.
\newblock {\em ACM Transactions on graphics (TOG)}, 35(4):1--11, 2016.

\bibitem{ezzat1998miketalk}
Tony Ezzat and Tomaso Poggio.
\newblock Miketalk: A talking facial display based on morphing visemes.
\newblock In {\em Proceedings Computer Animation'98 (Cat. No. 98EX169)}, pages
  96--102. IEEE, 1998.

\bibitem{ezzat2000visual}
Tony Ezzat and Tomaso Poggio.
\newblock Visual speech synthesis by morphing visemes.
\newblock {\em International Journal of Computer Vision}, 38:45--57, 2000.

\bibitem{fan2022faceformer}
Yingruo Fan, Zhaojiang Lin, Jun Saito, Wenping Wang, and Taku Komura.
\newblock Faceformer: Speech-driven 3d facial animation with transformers.
\newblock In {\em Proceedings of the IEEE/CVF Conference on Computer Vision and
  Pattern Recognition}, pages 18770--18780, 2022.

\bibitem{fanelli20103}
Gabriele Fanelli, Juergen Gall, Harald Romsdorfer, Thibaut Weise, and Luc
  Van~Gool.
\newblock A 3-d audio-visual corpus of affective communication.
\newblock {\em IEEE Transactions on Multimedia}, 12(6):591--598, 2010.

\bibitem{feng2021learning}
Yao Feng, Haiwen Feng, Michael~J Black, and Timo Bolkart.
\newblock Learning an animatable detailed 3d face model from in-the-wild
  images.
\newblock {\em ACM Transactions on Graphics (ToG)}, 40(4):1--13, 2021.

\bibitem{filntisis2023spectre}
Panagiotis~P Filntisis, George Retsinas, Foivos Paraperas-Papantoniou,
  Athanasios Katsamanis, Anastasios Roussos, and Petros Maragos.
\newblock Spectre: Visual speech-informed perceptual 3d facial expression
  reconstruction from videos.
\newblock In {\em Proceedings of the IEEE/CVF Conference on Computer Vision and
  Pattern Recognition}, pages 5744--5754, 2023.

\bibitem{guan2023stylesync}
Jiazhi Guan, Zhanwang Zhang, Hang Zhou, Tianshu Hu, Kaisiyuan Wang, Dongliang
  He, Haocheng Feng, Jingtuo Liu, Errui Ding, Ziwei Liu, et~al.
\newblock Stylesync: High-fidelity generalized and personalized lip sync in
  style-based generator.
\newblock In {\em Proceedings of the IEEE/CVF Conference on Computer Vision and
  Pattern Recognition}, pages 1505--1515, 2023.

\bibitem{guo2021ad}
Yudong Guo, Keyu Chen, Sen Liang, Yong-Jin Liu, Hujun Bao, and Juyong Zhang.
\newblock Ad-nerf: Audio driven neural radiance fields for talking head
  synthesis.
\newblock In {\em Proceedings of the IEEE/CVF International Conference on
  Computer Vision}, pages 5784--5794, 2021.

\bibitem{jang2016categorical}
Eric Jang, Shixiang Gu, and Ben Poole.
\newblock Categorical reparameterization with gumbel-softmax.
\newblock {\em arXiv preprint arXiv:1611.01144}, 2016.

\bibitem{kalberer2002speech}
Gregor~A Kalberer, Pascal M{\"u}ller, and Luc Van~Gool.
\newblock Speech animation using viseme space.
\newblock In {\em VMV}, pages 463--470, 2002.

\bibitem{kalberer2001face}
Gregor~A Kalberer and Luc Van~Gool.
\newblock Face animation based on observed 3d speech dynamics.
\newblock In {\em Proceedings Computer Animation 2001. Fourteenth Conference on
  Computer Animation (Cat. No. 01TH8596)}, pages 20--251. IEEE, 2001.

\bibitem{karras2017audio}
Tero Karras, Timo Aila, Samuli Laine, Antti Herva, and Jaakko Lehtinen.
\newblock Audio-driven facial animation by joint end-to-end learning of pose
  and emotion.
\newblock {\em ACM Transactions on Graphics (TOG)}, 36(4):1--12, 2017.

\bibitem{kingma2014adam}
Diederik~P Kingma and Jimmy Ba.
\newblock Adam: A method for stochastic optimization.
\newblock {\em arXiv preprint arXiv:1412.6980}, 2014.

\bibitem{lee2022autoregressive}
Doyup Lee, Chiheon Kim, Saehoon Kim, Minsu Cho, and Wook-Shin Han.
\newblock Autoregressive image generation using residual quantization.
\newblock In {\em Proceedings of the IEEE/CVF Conference on Computer Vision and
  Pattern Recognition}, pages 11523--11532, 2022.

\bibitem{massaro201212}
DW Massaro, MM Cohen, M Tabain, J Beskow, and R Clark.
\newblock Animated speech: research progress and applications.
\newblock 2012.

\bibitem{mittal2020animating}
Gaurav Mittal and Baoyuan Wang.
\newblock Animating face using disentangled audio representations.
\newblock In {\em Proceedings of the IEEE/CVF Winter Conference on Applications
  of Computer Vision}, pages 3290--3298, 2020.

\bibitem{ng2022learning}
Evonne Ng, Hanbyul Joo, Liwen Hu, Hao Li, Trevor Darrell, Angjoo Kanazawa, and
  Shiry Ginosar.
\newblock Learning to listen: Modeling non-deterministic dyadic facial motion.
\newblock In {\em Proceedings of the IEEE/CVF Conference on Computer Vision and
  Pattern Recognition}, pages 20395--20405, 2022.

\bibitem{oord2018representation}
Aaron van~den Oord, Yazhe Li, and Oriol Vinyals.
\newblock Representation learning with contrastive predictive coding.
\newblock {\em arXiv preprint arXiv:1807.03748}, 2018.

\bibitem{razavi2019generating}
Ali Razavi, Aaron Van~den Oord, and Oriol Vinyals.
\newblock Generating diverse high-fidelity images with vq-vae-2.
\newblock {\em Advances in neural information processing systems}, 32, 2019.

\bibitem{richard2021meshtalk}
Alexander Richard, Michael Zollh{\"o}fer, Yandong Wen, Fernando De~la Torre,
  and Yaser Sheikh.
\newblock Meshtalk: 3d face animation from speech using cross-modality
  disentanglement.
\newblock In {\em Proceedings of the IEEE/CVF International Conference on
  Computer Vision}, pages 1173--1182, 2021.

\bibitem{taylor2017deep}
Sarah Taylor, Taehwan Kim, Yisong Yue, Moshe Mahler, James Krahe,
  Anastasio~Garcia Rodriguez, Jessica Hodgins, and Iain Matthews.
\newblock A deep learning approach for generalized speech animation.
\newblock {\em ACM Transactions On Graphics (TOG)}, 36(4):1--11, 2017.

\bibitem{taylor2012dynamic}
Sarah~L Taylor, Moshe Mahler, Barry-John Theobald, and Iain Matthews.
\newblock Dynamic units of visual speech.
\newblock In {\em Proceedings of the 11th ACM SIGGRAPH/Eurographics conference
  on Computer Animation}, pages 275--284, 2012.

\bibitem{van2017neural}
Aaron Van Den~Oord, Oriol Vinyals, et~al.
\newblock Neural discrete representation learning.
\newblock {\em Advances in neural information processing systems}, 30, 2017.

\bibitem{verma2003using}
Ashish Verma, Nitendra Rajput, and L~Venkata Subramaniam.
\newblock Using viseme based acoustic models for speech driven lip synthesis.
\newblock In {\em 2003 IEEE International Conference on Acoustics, Speech, and
  Signal Processing, 2003. Proceedings.(ICASSP'03).}, volume~5, pages V--720.
  IEEE, 2003.

\bibitem{xing2023codetalker}
Jinbo Xing, Menghan Xia, Yuechen Zhang, Xiaodong Cun, Jue Wang, and Tien-Tsin
  Wong.
\newblock Codetalker: Speech-driven 3d facial animation with discrete motion
  prior.
\newblock In {\em Proceedings of the IEEE/CVF Conference on Computer Vision and
  Pattern Recognition}, pages 12780--12790, 2023.

\bibitem{xu2013practical}
Yuyu Xu, Andrew~W Feng, Stacy Marsella, and Ari Shapiro.
\newblock A practical and configurable lip sync method for games.
\newblock In {\em Proceedings of Motion on Games}, pages 131--140. 2013.

\bibitem{zhang2021facial}
Chenxu Zhang, Yifan Zhao, Yifei Huang, Ming Zeng, Saifeng Ni, Madhukar
  Budagavi, and Xiaohu Guo.
\newblock Facial: Synthesizing dynamic talking face with implicit attribute
  learning.
\newblock In {\em Proceedings of the IEEE/CVF international conference on
  computer vision}, pages 3867--3876, 2021.

\bibitem{zhou2018visemenet}
Yang Zhou, Zhan Xu, Chris Landreth, Evangelos Kalogerakis, Subhransu Maji, and
  Karan Singh.
\newblock Visemenet: Audio-driven animator-centric speech animation.
\newblock {\em ACM Transactions on Graphics (TOG)}, 37(4):1--10, 2018.

\end{thebibliography}
}
\newpage
\begin{strip}
\centering
\Large \textbf{\emph{Supplementary Materials} \\
Probabilistic Speech-Driven 3D Facial Motion Synthesis:\\ New Benchmarks, Methods, and Applications}
\end{strip}

\appendix
%%%%%%%%% BODY TEXT

\section{Supplemental Video}
Please see the \textbf{accompanying video} for an overview of our work and qualitative examples from our model.

\section{Overview}

The following sections provide additional methodology and/or results that were not included in the main paper. 

Section \ref{sec:pretrained-metrics} provides more details on the metrics, including (i) limitations of maximal lip vertex error for evaluating probabilistic models and (ii) details of the pretrained models used for evaluation.

Section \ref{sec:supp-results} provides the complete table of benchmark results corresponding to Figure 2 of the main paper, a discussion of CodeTalker \cite{xing2023codetalker}, and additional ablation results for our model.

Section \ref{sec:efficiency} discusses \emph{efficiency} of our method, including (i) a knowledge distillation strategy for amortizing the sampling strategies, and (ii) diversity \vs efficiency trade-off that can be achieved by sampling fewer codes from our auto-regressive model at inference time.

Section \ref{sec:implementation} provides implementation and training details that were deferred from the main text.

Section \ref{sec:discussion} discusses the limitations and ethical considerations of this work.

\begin{figure}[t]
\centering
\includegraphics[scale=0.35]{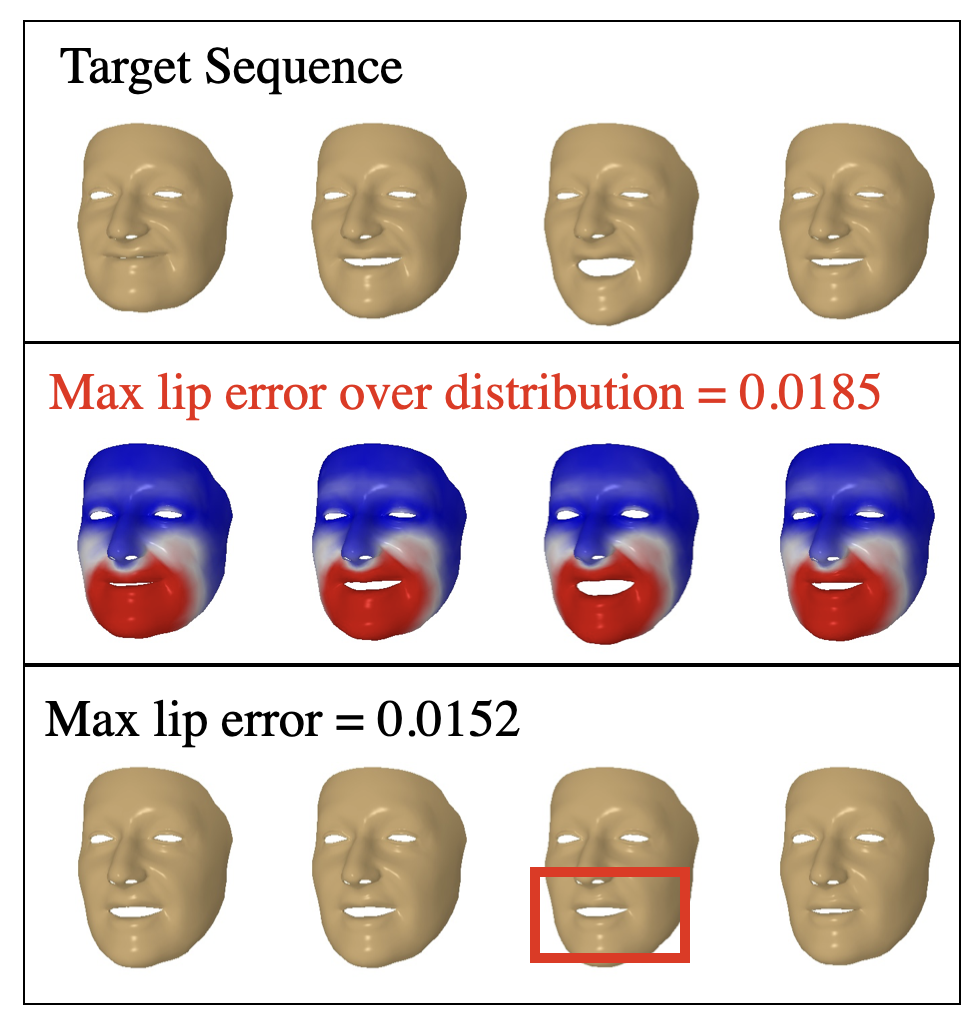}
\caption{\textbf{Limitations of $\ell_{vertex}$ as a metric.} Probabilistic models (row 2) generate 3D facial motions with diversity, as shown by the color map of standard deviation. They may achieve worse $\ell_{vertex}$ compared to deterministic models (row 3), despite being able to generate a mesh sequence that matches the ground truth sequence better (row 1). See text for details.}
\label{fig:lip_vertex_error}
\end{figure}

\section{Metrics} \label{sec:pretrained-metrics}

\subsection{Limitation of Maximal Lip Vertex Error}
Maximal lip vertex error ($\ell_{vertex}$) is a metric that measures the maximum difference in lip vertices between the ground truth mesh and a mesh generated from the model. This metric is used as a proxy for lip articulation quality in existing works, but as a standalone metric, it has limitations for evaluating probabilistic models. As shown in Supplemental Figure \ref{fig:lip_vertex_error}, a probabilistic model (row 2) can generate lip articulation that is more similar to the ground truth (row 1), but due to variations between samples, have larger $\ell_{vertex}$ compared to a deterministic model (row 3) that generates an over-averaged result. Our proposed lip vertex metrics, $\ell_{cover}$ and $\ell_{mean}$, address this limitation and provide a more complete picture of performance. Overall, there is a need to look across multiple metrics (sync score, FD score, lip vertex error) when evaluating speech-driven 3D facial motion synthesis.

\subsection{Audio-Mesh Synchronization Networks}
\begin{figure}[t]
\centering
\includegraphics[scale=0.5]{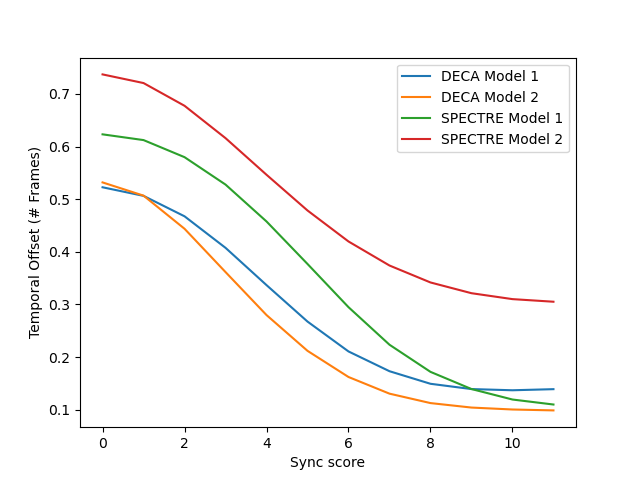}
\caption{\textbf{SyncNet Evaluation} All of our synchronization networks can detect individual frames of temporal shift between audio and 3D facial mesh sequences.}
\label{fig:syncnet}
\end{figure}

The audio-mesh synchronization networks are trained using InfoNCE contrastive loss \cite{oord2018representation} with a batch size of $64$, \emph{i.e.,} for each 3D mesh sequence, we sample $63$ negative audio examples that are either semantically misaligned (taken from a different clip) or temporally misaligned (taken from a different time point of the same clip). Supplemental Figure \ref{fig:syncnet} shows plots of the models evaluated on held-out ground truth audio-mesh pairs with increasing temporal misalignment. The results indicate that all the pretrained networks are sensitive to individual frames of audio-mesh misalignment.

\subsection{Speaking Style Recognition Network} The style recognizer is trained on 3D facial motion (deformation between animated and neutral face meshes). The facial motion encoder uses a similar architecture as our RVQ encoder with standard 1D convolutional blocks instead of causal 1D convolutional blocks. We use angular margin loss \cite{deng2019arcface} to maximize the cosine similarity between embeddings from the same speaker while minimizing cosine similarity with other speakers. The performance of the models for DECA and SPECTRE are shown in the GT line of Table 2 in the main paper.

\section{Additional Results}\label{sec:supp-results}
\subsection{Complete Table - Main Figure 2}
Supplemental Table \ref{table:complete_table} shows the full results corresponding to Figure 2 of the main paper. Our approach outperforms the existing methods across the board. Importantly, while deterministic methods (\emph{i.e.,} Faceformer+Style) achieve good lip synchronization and lip vertex error, they suffer in diversity/realism as highlighted in \textcolor{red}{red}.

\paragraph{Sync score}
Existing deterministic methods (FaceFormer, Faceformer+Style) achieve better synchronization than existing probabilistic methods (Meshtalk, MeshTalk-ND, MeshTalk+Style, MeshTalk-ND+Style). Our probabilistic method achieves the highest sync scores out of all methods, particularly when we use our sampling strategies to trade off diversity for greater speech fidelity (Ours+Avg100). 

\paragraph{Frechet distance} Deterministic methods (VOCA, Faceformer, Faceformer+Style) suffer on this metric, as highlighted in \textcolor{red}{red}, suggesting that the generated facial motions are unrealistic. The probabilistic methods perform better on this metric, with our method outperforming MeshTalk on three out of four cases.

\paragraph{Maximal Lip Vertex Error} Deterministic methods (VOCA, Faceformer, Faceformer+Style) achieve lower $\ell_{vertex}$, which measures the maximum vertex error between the ground truth and one synthesized mesh sequence. However, this does not take into account the diversity of probabilistic methods. When we compute the maximal vertex error between the ground truth and the average of many synthesized sequences ($\ell_{mean}$), our approach matches Faceformer+Style and outperforms the others. We also achieve the lowest coverage error ($\ell_{cover}$), suggesting that the ground truth sequences are closest to our sampling distribution. Finally, when we use sampling strategies (Ours+Avg100), we are able to trade off the coverage of our model ($\ell_{cover}$) for improved precision ($\ell_{vertex}$).

\begin{table*}[t]
    \setlength{\tabcolsep}{0.3em} % increase space between columns slightly for easier reading
    \newcommand{\indentrow}{\hspace{6pt}}
    
    \centering
    \footnotesize
    \resizebox{1.8\columnwidth}{!}{  % stretch/shrink table to be exactly columnwidth
        \begin{tabular}{l|c|c|c|c|c|c|c|}
            \toprule
            \multicolumn{1}{l|}{\textbf{Model}} & \multicolumn{2}{c|}{\textbf{Sync score} $\uparrow$} & \multicolumn{2}{c|}{\textbf{Frechet distance} $\downarrow$} & \multicolumn{3}{c|}{\textbf{Maximal Lip Vertex Error ($\times 10^{-3}$)} $\downarrow$} \\
            \midrule
            \textbf{DECA} & \textbf{Model 1}  & \textbf{Model 2} & \textbf{Model 1} & \textbf{Model 2} $\downarrow$ & \textbf{$\mathbf{\ell}_{vertex}$}  & \textbf{$\mathbf{\ell}_{cover}$}  & \textbf{$\mathbf{\ell}_{mean}$}  \\
            \midrule
            VOCA & 0.137 & 0.271 & \textcolor{red}{22.0} & \textcolor{red}{2.94} & 10.0 & 10.0 & 10.0 \\
            FaceFormer & 0.348 & 0.361 & \textcolor{red}{13.9} & \textcolor{red}{2.44} & 9.9 & 9.9 & 9.9 \\
            MeshTalk & 0.262 & 0.174 &5.2 & 0.48 & 11.1 & 6.9 &  10.3\\
            MeshTalk-ND & 0.286 & 0.284 & 1.3 & 0.40 & 13.2 &  8.6 & 10.5 \\
            \midrule 
            FaceFormer+Style & 0.369 & 0.441 & \textcolor{red}{13.2} & \textcolor{red}{1.89} & \textbf{7.9} & 7.9 & \textbf{7.9} \\
            MeshTalk+Style & 0.286 & 0.203 & 4.7 & 0.64 & 8.4& \underline{6.3}& \underline{8.1} \\
            MeshTalk-ND+Style & 0.298 & 0.302 &\underline{1.0} & \underline{0.34}& 11.9 & 7.6 & 9.5 \\
            \midrule
            Ours & \underline{0.463} & \underline{0.464} & \textbf{0.9} & \textbf{0.23} & 10.8& \textbf{6.0} & \textbf{7.9} \\
            Ours+Avg100 & \textbf{0.684} & \textbf{0.600} & 7.1 & 1.18 & \underline{8.3} & 7.1 & 8.2 \\
            \midrule
            \textbf{SPECTRE} & \textbf{Model 1}  & \textbf{Model 2} & \textbf{Model 1} & \textbf{Model 2} $\downarrow$ & \textbf{$\mathbf{\ell}_{vertex}$}  & \textbf{$\mathbf{\ell}_{cover}$}  & \textbf{$\mathbf{\ell}_{mean}$}  \\
            \midrule
            VOCA & 0.357 & 0.290 & \textcolor{red}{524.9} & \textcolor{red}{66.8} & 15.5& 15.5& 15.5\\
            FaceFormer & 0.393 & 0.423 & \textcolor{red}{449.5} & \textcolor{red}{70.4} & 15.6 & 15.6 & 15.6 \\
            MeshTalk & 0.309 & 0.302 & 227.9 & 37.3 & 17.7 &12.4&16.1 \\
            MeshTalk-ND & 0.327 & 0.436 &49.1 & \underline{7.4} & 20.3 & 13.1 & 16.0 \\
            \midrule 
            FaceFormer+Style & 0.438 & \underline{0.576} & \textcolor{red}{351.0} & \textcolor{red}{40.5} & \textbf{12.8}& 12.8& \textbf{12.8}\\
            MeshTalk+Style & 0.331 & 0.372 &178.5 & 12.7 & \underline{13.9} & \underline{10.2} & 13.2\\
            MeshTalk-ND+Style & 0.325 & 0.474 & \underline{43.2} & \textbf{6.9} & 17.9 & 11.8 & 14.5 \\
            \midrule
            Ours & \underline{0.444} & 0.520 & \textbf{40.9} & 8.4 & 18.2 & \textbf{10.0} & \underline{13.0} \\
            Ours+Avg100 & \textbf{0.565} & \textbf{0.591} & 199.3 & 30.3 & \underline{13.9} &11.9 & 13.7 \\
            \bottomrule   
        \end{tabular}
    }
    \caption{\textbf{Benchmark Results} corresponding to Figure 2 in the main paper. Best results in each column are \textbf{bolded}, while second best results are \underline{underlined}. $\ell_{vertex}$, $\ell_{cover}$, and $\ell_{mean}$ denote the maximal lip vertex error, coverage error, and mean estimate error respectively and are computed with $|\mathcal{S}|=100$. See Section 4.2 of the main text for descriptions of the metrics. }
    \label{table:complete_table}
\end{table*}

\subsection{Discussion of CodeTalker}
CodeTalker \cite{xing2023codetalker} extends Faceformer \cite{fan2022faceformer} using a vector-quantized (VQ) autoencoder to learn a discrete 3D facial motion prior. While Faceformer uses an auto-regressive transformer to directly regress 3D mesh deformations, CodeTalker uses an auto-regressive transformer to regress the embeddings of the ground truth meshes in the latent space. Their training loss consists of a combination of regression errors over the embeddings and the original 3D mesh deformations after decoding, and training occurs in a teacher-forcing manner. During inference, the predicted embeddings are projected to the nearest codes in the VQ codebook before being decoded to produce 3D facial motion. The motivation is that the projection to the VQ codebook selects a mode in the distribution of 3D facial motions, whereas Faceformer regresses to the conditional mean of motion and produces over-smoothed outputs that do not correspond to any mode. Importantly, while their auto-regressive model selects codes from a pretrained codebook, it is deterministic and selects a code that is nearest to the regressed latent embedding.

We trained the original implementation of CodeTalker by the authors on our data, as well as our own re-implementation using our RVQ codebook and auto-regressive architecture. While training the VQ codebook produced good reconstructions of 3D facial motion, we found that training the auto-regressive model using the combination of regression losses failed to converge to a reasonable result on our data. This is likely due to the large scale and diversity of our dataset compared to VocaSet \cite{cudeiro2019capture} and BIWI \cite{fanelli20103}, which leads to a high-variance, multi-modal distribution in the latent space that is difficult to regress. 

While we are unable to converge to a reasonable result with their original loss, we note that conceptually, taking the \emph{expectation} of the code embeddings sampled from our model at each time point would produce an equivalent result to performing regression in the latent space. In other words, the expected output of CodeTalker can be achieved by performing code averaging as in Section 3.3 with an infinite number of codes. Therefore, we expect the performance of CodeTalker to be the limiting case of the trend of the green points in Figure 2(a-b) of the main text.

\subsection{Additional Ablations}

\begin{table}[t]
    \setlength{\tabcolsep}{0.3em} % increase space between columns slightly for easier reading
    \newcommand{\indentrow}{\hspace{6pt}}
    
    \centering
    \footnotesize
    \resizebox{1\columnwidth}{!}{  % stretch/shrink table to be exactly columnwidth
        \begin{tabular}{ll|cc}
            \toprule
            Temporal Model & Uses Ref. Style? & Sync Score $\uparrow$ & Averaged FD $\downarrow$ \\
            \midrule
            Transformer block & no & \textbf{0.45} & \textbf{0.51} \\
            Ours & no & 0.43 & 0.73 \\
            \midrule
            Transformer block & yes & 0.43 & \textbf{0.36} \\ 
            Ours & yes & \textbf{0.44} & 0.54 \\

            \bottomrule   
        \end{tabular}
    }
    \caption{\textbf{Choice of Temporal Model} Comparison of transformer \vs masked convolution blocks for the temporal model on the DECA meshes. See text for details.}
    \label{table:temporal_model}
\end{table}

\begin{table}[t]
    \setlength{\tabcolsep}{0.3em} % increase space between columns slightly for easier reading
    \newcommand{\indentrow}{\hspace{6pt}}
    
    \centering
    \footnotesize
    \resizebox{1\columnwidth}{!}{  % stretch/shrink table to be exactly columnwidth
        \begin{tabular}{l|cc}
            \toprule
            Audio Encoder & First Code CE $\downarrow$ & Average Code CE $\downarrow$ \\
            \midrule
            Wav2Vec 2.0 \cite{baevski2020wav2vec} & 2.09 & 2.54  \\
            Ours (trained from scratch) & \textbf{1.97} & \textbf{2.50} \\
            \bottomrule   
        \end{tabular}
    }
    \caption{\textbf{Choice of Audio Encoder} Comparison of Wav2Vec 2.0 \cite{baevski2020wav2vec} and our audio encoder trained on scratch on the DECA meshes. CE: cross-entropy loss on held-out set (lower is better). See text for details.}
    \label{table:audio_encoder}
\end{table}

\paragraph{Choice of Temporal Model} We show preliminary results of our model trained with different temporal models in Supplemental Table \ref{table:temporal_model}. We found that using a transformer as the temporal model in the absence of a reference style clip improves both the synchronization and realism/diversity of the outputs, as measured by sync score and Frechet distance respectively. However, the results were more varied when we provide additional information through a reference style clip. Use of a transformer for the temporal model may improve diversity at the cost of synchronization.

\paragraph{Choice of Audio Encoder} Several recent works, namely Faceformer \cite{fan2022faceformer} and CodeTalker \cite{xing2023codetalker} use a self-supervised and pretrained wav2vec 2.0 speech model \cite{baevski2020wav2vec} as the audio encoder. While this may prevent overfitting of the audio encoder on small datasets as VocaSet \cite{cudeiro2019capture} and BIWI \cite{fanelli20103}, we found that using a pretrained audio encoder was not necessary for a large-scale dataset like VoxCeleb2 \cite{chung2018voxceleb2}. As shown in Supplemental Table \ref{table:audio_encoder}, in our preliminary experiences, we found that using the pretrained speech model did not improve results.

\section{Improving Efficiency}\label{sec:efficiency}
While the focus of the methodology and results in the main paper was primarily on the quality and diversity of the model, for certain applications (\emph{e.g.,} real-time speech-driven 3D avatars), the efficiency of the method is also important. In this section, we elaborate on improving the speed/efficiency of our method. 

\subsection{Knowledge Distillation}  In Section 3.3 of the main text, we described sampling strategies for trading off the diversity of the auto-regressive model for improved precision and fidelity. However, this strategy increases the inference speed, as multiple codes need to be sampled and aggregated. We propose a knowledge distillation strategy for amortizing this added sampling time. Recall that ${\mathbf{j}}$ denotes a matrix of codebook indices indexed by time $t$ and depth $d$ corresponding to the real facial motion $\mathbf{x}$. We obtain the new targets $\hat{\mathbf{j}}$ for the student by:
\begin{enumerate}
    \item Computing the audio-visual context using the temporal model $\mathbf{h}_{av}[t]$ in a teacher-forcing manner, \emph{i.e.,} inputting the ground truth ${\mathbf{j}}$ into Equation (2) in the main text.
    \item Sampling codes from the depth model using Equation (3) without teacher-forcing. Namely, we use $\mathbf{h}_{av}[t]$ from Step 1 to compute $v_{t2}$, but use the sampled codes in place of the ground truth codes for computing $v_{td}, d \geq 3$.
    \item Aggregating the sampled codes using strategies discussed in Section 3.3 of the main text and reprojecting them to the RVQ codebook to obtain new indices $\hat{\mathbf{j}}$.
\end{enumerate} 
The student model is trained in a teacher-forcing manner using both the ground truth codes ${\mathbf{j}}$ as well as the new targets $\hat{\mathbf{j}}$. Specifically, we use ${\mathbf{j}}$ as input to the temporal model, and we use $\hat{\mathbf{j}}$ as input to the depth model. We also use $\hat{\mathbf{j}}$ as the targets for optimizing the student model. As shown in Supplemental Figure \ref{fig:efficiency}, this enables us to distill the sampled and aggregated labels from a teacher model (blue) to a student model (yellow, `16') with improved inference time.

\begin{figure}[t]
\centering
\includegraphics[scale=0.3]{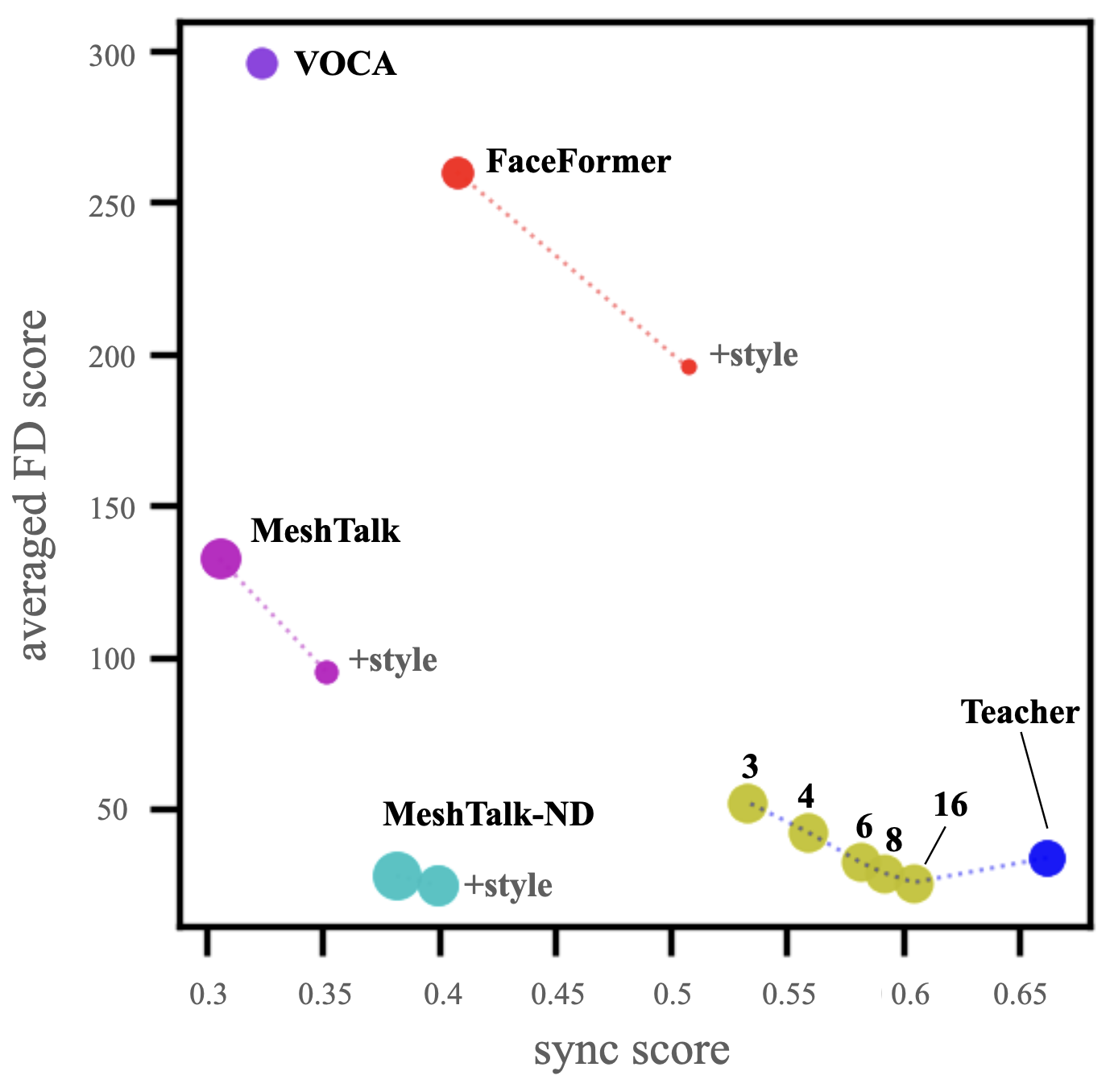}
\caption{\textbf{Improving Model Efficiency} Results are shown for the SPECTRE meshes. See text for details. }
\label{fig:efficiency}
\end{figure}

\subsection{Quality \vs Efficiency Trade-off}
One of the advantages of the coarse-to-fine design of the RVQ codebook is the possibility of improving efficiency by predicting and decoding codes. We show that this can be done while still achieving high synchronization for as few as 3 codes (out of depth of 16). As shown by the yellow points in Supplemental Figure \ref{fig:efficiency}, reducing the number of codes yields a \textbf{quality \vs efficiency trade-off}, where we can achieve improved speed/efficiency at the cost of losing finer 3D motions. 

\section{Implementation Details}\label{sec:implementation}

\paragraph{RVQ Autoencoder}
The 3D facial motion encoder and decoder consist of 1D convolutional blocks. The inputs and outputs are 3D facial motion represented by mesh vertex deformations, \emph{i.e.}, the difference between the mesh vertex positions for animated and neutral expressions. The encoder consists of a 1D convolutional layer with kernel size of $1$ to aggregate information over mesh vertex deformations, then two 1D causal convolutional layers with kernel size of $3$ to aggregate information over time. The decoder consists of the same blocks in the reverse order. For input size $\mathbf{x} \in \mathbb{R}^{T \times 3V}$, the size of the latent embeddings is $Z \in \mathbb{R}^{T \times N_C}$, where $N_C$ is the dimensionality of the codes in codebook $\mathcal{C}$. In practice, we use a shared codebook \cite{lee2022autoregressive} with $D=16$, $|\mathcal{C}|=256$ and $N_C = 128$.

\paragraph{Audio encoder} Following \cite{richard2021meshtalk}, our audio encoder consists of 1D convolutional blocks operating over mel-spectrograms of 1s audio samples centered at each visual frame. 

\paragraph{Reference Clip Encoder} The reference clip is encoded using the same architecture as the RVQ encoder, except standard convolutional layers are used in place of the causal convolutions. 

\paragraph{Two-Stage Auto-Regressive Model} The temporal auto-regressive model consists of four masked causal convolutional layers \cite{richard2021meshtalk} with kernel size of $2$ and increasing dilation of $1,2,4,8$ for gathering audio-visual context. The depth auto-regressive model consists of a masked transformer self-attention block with embedding size of $64$. 

\paragraph{Sampling Strategies} For KNN-based sampling, we use $N=100$ and $K=3$. For SyncNet-based sampling, we take the top $1/2$ codes based on the synchronization score. For code averaging, we vary the number of codes averaged depending on the desired diversity vs. fidelity trade-off.

\paragraph{Training} We train the RVQ autoencoder and two-stage auto-regressive model for approximately $150$ and $200$ epochs respectively with Adam optimizer \cite{kingma2014adam} with learning rate of $10^{-4}$. For knowledge distillation, we train the student for approximately $100$ epochs. The two-stage auto-regressive model is trained in a teacher-forcing manner. We use both stochastic sampling and soft code targets \cite{lee2022autoregressive}.

\section{Additional Discussion}\label{sec:discussion}
\paragraph{Limitations.} (1) Our benchmark dataset relies on state-of-the-art monocular face reconstruction techniques \cite{feng2021learning, filntisis2023spectre} and the VoxCeleb2 video dataset \cite{chung2018voxceleb2}. The quality of the face meshes is limited compared to those reconstructed from high-resolution multi-view videos. (2) While our model can achieve real-time synthesis on high-end GPUs, it does not run in real-time on standard consumer-grade hardware. We leave improvements along these directions to future work.

\paragraph{Ethical Considerations.} The datasets and models used in this work are intended for research purposes only. While meshes from this work can be used to render photo-realistic content, they should not be used to generate videos of individuals without their consent.

%%%%%%%%% REFERENCES

\end{document}